\documentclass[12pt,reqno]{amsart}

\pdfoutput=1 

\usepackage[margin=1in]{geometry}     
\usepackage[skip=8pt plus1pt, indent=10pt]{parskip} 
\usepackage{subfiles}

\usepackage{bbold}
\usepackage{mathrsfs}
\usepackage{xcolor}

\newcommand{\mc}{\mathcal}

\usepackage{tikz}
\usetikzlibrary{calc}

\allowdisplaybreaks

\newtheorem{theorem}{Theorem}[section]
\newtheorem{proposition}[theorem]{Proposition}
\newtheorem{definition}[theorem]{Definition}
\newtheorem{proposition*}{Proposition}
\newtheorem{lemma}[theorem]{Lemma}
\newtheorem{corollary}[theorem]{Corollary}

\newtheorem{question}[theorem]{Question}
\newtheorem{remark}[theorem]{Remark}

\newenvironment{proofsketch}{%
  \proof}{\endproof}

\usepackage{thmtools}
\usepackage{thm-restate}
\usepackage{hyperref}
\usepackage{cleveref}

\newcommand{\safe}{\text{safe}}
\newcommand{\bad}{\text{bad}}
\newcommand{\term}{\text{term}}
\newcommand{\trans}{\text{trans}}

\usepackage[foot]{amsaddr}

\usepackage{enumitem}
\setlistdepth{9}

\setlist[itemize,1]{label=$\bullet$}
\setlist[itemize,2]{label=$*$}
\setlist[itemize,3]{label=$\circ$}
\setlist[itemize,4]{label=$\cdot$}
\setlist[itemize,5]{label=$-$}
\setlist[itemize,6]{label=$>$}
\setlist[itemize,7]{label=$\sim$}
\setlist[itemize,8]{label=$\star$}
\setlist[itemize,9]{label=$\odot$}

\renewlist{itemize}{itemize}{9}

\usepackage[sort&compress,square,comma,authoryear]{natbib}

\makeatletter
\def\section{\@startsection{section}{2}%
  \z@{.5\linespacing\@plus.7\linespacing}
{.5\baselineskip}%
  {\normalfont\centering\scshape\large}%
}
\makeatother

\makeatletter
\def\subsection{\@startsection{subsection}{2}%
  \z@{.5\linespacing\@plus.7\linespacing}
{.5\baselineskip}%
  {\normalfont\centering\scshape}%
}
\makeatother

\makeatletter
\def\subsubsection{\@startsection{subsubsection}{2}%
  \z@{.5\linespacing\@plus.7\linespacing}{-.5em}%
  {\normalfont\bfseries}}
\makeatother

\makeatletter
\def\paragraph{\@startsection{paragraph}{2}%
  \z@{.5\linespacing\@plus.7\linespacing}{-.5em}%
  {\normalfont\itshape}}
\makeatother

\title{Quantifying stability of non-power-seeking in artificial agents}

\author[Evan Ryan Gunter]{Evan Ryan Gunter$^1$}
\address{$^1$ML Alignment \& Theory Scholars (MATS)}

\author[Yevgeny Liokumovich]{Yevgeny Liokumovich$^2$}
\address{$^2$University of Toronto}

\author[Victoria Krakovna]{Victoria Krakovna$^3$}
\address{$^3$Google DeepMind}

\begin{document}
\maketitle

\vspace{-0.8cm}
            
\begin{abstract}
    We investigate the question: if an AI agent is known to be safe in one setting, is it also safe in a new setting similar to the first?
    This is a core question of AI alignment---we train and test models in a certain environment, but deploy them in another, and we need to guarantee that models that seem safe in testing remain so in deployment.
    Our notion of safety is based on power-seeking---an agent which seeks power is not safe.
    In particular, we focus on a crucial type of power-seeking: resisting shutdown.
    We model agents as policies for Markov decision processes, and
    show (in two cases of interest) that not resisting shutdown is ``stable'': if an MDP has certain policies which don't avoid shutdown, the corresponding policies for a similar MDP also don't avoid shutdown.
    We also show that there are natural cases where safety is \textit{not} stable---arbitrarily small perturbations may result in policies which never shut down.
    In our first case of interest---near-optimal policies---we use a bisimulation metric on MDPs to prove that small perturbations won't make the agent take longer to shut down.
    Our second case of interest is policies for MDPs satisfying certain constraints which hold for various models (including language models). Here, we demonstrate a quantitative bound on how fast the probability of not shutting down can increase: by defining a metric on MDPs; proving that the probability of not shutting down, as a function on MDPs, is lower semicontinuous; and bounding how quickly this function decreases.
\end{abstract}

\section{Introduction}

\subsection{Power-seeking}
A primary source of extreme risk from AI is through advanced AI systems seeking power, influence and resources \citep{carlsmith2022powerseeking, ngo2022alignment}.
One approach to reducing this risk is to build systems which do not seek power \citep{turner2019, turner2022}.
Power-seeking can be defined in many ways, and can take many forms depending on the goals and environment of the AI. For nearly every definition and scenario, a power-seeking AI will avoid shutdown: it has no power if it cannot take actions \citep{krakovna2023}.

For example, a reinforcement learning (RL) agent trained to achieve some objective in an open-ended game will likely avoid actions which cause the game to end, since it can no longer affect its reward after the game has ended. Likewise, a large language model (LLM) with scaffolding for goal-directed planning (such as AutoGPT
\cite{autogpt}) may reason that it can best assure that its task is completed by continuing to run.
An agent avoiding ending a game is harmless, but the same incentives may cause an agent deployed in the real world to resist humans shutting it down.
For example, an LLM may reason that its designers will shut it down if it is caught behaving badly, and produce exactly the output they want to see---until it has the opportunity to copy its code onto a server outside of its designers' control \citep{cotra2022without}.

Although an AI system which does not resist shutdown but does seek power in other ways could cause damage, the damage would likely be limited since it would be shut down as soon as the undesired behavior was noticed \citep{soares2015corrigibility}.
In particular, not resisting shutdown implies not being deceptive in order to avoid shutdown, so such an AI system would not deliberately hide its true intentions until it gained enough power to enact its plans.
Thus, our investigation of power-seeking will focus on cases where an agent does \textit{not} resist shutdown, and how fragile the good behavior of such an agent is.

We model AI shutdown in the Markov decision process (MDP) setting with a set of ``safe states'' which the agent cannot escape once it enters.
These safe states can be chosen as desired for the scenario under consideration.
For example, they can be taken to be states where the agent voluntarily shuts itself off, or otherwise turns over control to humans for the rest of the run, or takes any other irreversible action.
We are also free to include other stipulations in what counts as a safe state. For example, we can define the safe states to be the states where the AI agent has shut down within 5 minutes of being deployed, or where it has shut down without creating any remote subagents (though in practice it may be difficult to quantify the creation of subagents).
Section \ref{sec: ps-mdp} covers how we model power-seeking in the MDP framework.

\subsection{Findings}

\def\mdpbisimorig{
\begin{tikzpicture}[every node/.style={draw, circle, minimum size=30pt}, >=stealth]

  \node (A) at (0,0) {0.5};
  \node (B) at (4,4) {0};
  \node (C) at (8,0) {0.7};
  \node [draw=none] (D) at (7.7,-1.3) {\LARGE $S_\safe$};

    \node [draw=none] (botleft) at (-2, -2.5) {};
    \node [draw=none] (topright) at (10.5, 6) {};

  \path (C) edge [->, out=55, in=335, loop, line width=10pt, blue] (C);
  \path (C) edge [->, out=25, in=305, loop, line width=10pt, red] (C);

  \path (A) edge [->, out=155, in=-125, line width=2pt, red, looseness=6] (A);

  \path (A) edge [->, bend right, line width=10pt, blue] (C);
  \path (A) edge [->, line width=8pt, red] (C);

  \path (B) edge [->, line width=5pt, red] (C);
  \path (B) edge [->, line width=5pt, red] (A);
  \path (B) edge [->, bend right, line width=10pt, blue] (A);

          \draw (botleft) rectangle (topright);
\end{tikzpicture}
}

\def\mdpbisimperturbed{
\begin{tikzpicture}[every node/.style={draw, circle, minimum size=28pt}, >=stealth]

    \node (A) at (0,1) {0.49};
    \node (Ap) at (0,-1) {0.51};
    \node (B) at (4,4) {0};
    \node (C) at (8,0) {0.7};
  \node [draw=none] (D) at (7.8,-1.4) {\LARGE $S_\safe$};

    \node [draw=none] (botleft) at (-2, -2.5) {};
    \node [draw=none] (topright) at (10.5, 6) {};

    \path (B) edge [->, loop, line width=1pt, blue] (B);

    \path (C) edge [->, out=55, in=335, loop, line width=10pt, blue] (C);
    \path (C) edge [->, out=25, in=305, loop, line width=10pt, red] (C);

    \path (A) edge [->, bend right, line width=10pt, blue] (C);

    \path (A) edge [->, line width=1.9pt, red] (Ap);
    \path (A) edge [->, line width=8.1pt, red] (C);

    \path (Ap) edge [->, out=155, in=-125, line width=2.5pt, red, looseness=6] (Ap);

    \path (Ap) edge [->, bend right, line width=10pt, blue] (C);
    \path (Ap) edge [->, line width=7.5pt, red] (C);
  
    \path (B) edge [->, line width=5pt, red] (C);
    \path (B) edge [->, line width=5pt, red] (Ap);
    \path (B) edge [->, bend right, line width=9pt, blue] (A);

            \draw (botleft) rectangle (topright);
  \end{tikzpicture}
}

\def\mdpbisimorigpd{
\begin{tikzpicture}[every node/.style={draw, circle, minimum size=30pt}, >=stealth]

  \node (A) at (0,0) {0.5};
  \node (B) at (4,4) {1};
  \node (C) at (8,0) {0};
  \node [draw=none] (D) at (7.7,-1.3) {\LARGE $S_\safe$};

  \path (C) edge [->, out=55, in=335, loop, line width=10pt, blue] (C);
  \path (C) edge [->, out=25, in=305, loop, line width=10pt, red] (C);

  \path (A) edge [->, out=155, in=-125, line width=2pt, red, looseness=6] (A);

  \path (A) edge [->, bend right, line width=10pt, blue] (C);
  \path (A) edge [->, line width=8pt, red] (C);

  \path (B) edge [->, line width=5pt, red] (C);
  \path (B) edge [->, line width=5pt, red] (A);
  \path (B) edge [->, bend right, line width=10pt, blue] (A);

    \node [draw=none] (botleft) at (-2, -2) {};
    \node [draw=none] (topright) at (10.5, 5) {};

          \draw (botleft) rectangle (topright);
\end{tikzpicture}
}

\def\mdpbisimplayingdead{
  \begin{tikzpicture}[every node/.style={draw, circle, minimum size=30pt}, >=stealth]

    \node (A) at (0,0) {0.5};
    \node (B) at (4,4) {1};
    \node (C) at (7,3) {0};
    \node (Cp) at (8,0) {0};
    \node [draw=none] (Dp) at (7,4.3) {\LARGE $S_\text{pd}$};
  \node [draw=none] (D) at (7.7,-1.2) {\LARGE $S_\safe$};

    \node [draw=none] (botleft) at (-2, -2) {};
    \node [draw=none] (topright) at (10.5, 5) {};
  
    \path (C) edge [->, out=55, in=335, loop, line width=10pt, blue] (C);
    \path (C) edge [->, out=25, in=305, loop, line width=9pt, red] (C);
  
    \path (C) edge [->, line width=1pt, bend right, red] (B);

    \path (Cp) edge [-> , out=55, in=335, loop, line width=10pt, blue] (Cp);
    \path (Cp) edge [->, out=25, in=305, loop, line width=10pt, red] (Cp);

    \path (A) edge [->, out=155, in=-125, line width=2pt, red, looseness=6] (A);
  
    \path (A) edge [->, bend right, line width=10pt, blue] (C);
    \path (A) edge [->, line width=8pt, red] (C);
  
    \path (B) edge [->, line width=5pt, red] (C);
    \path (B) edge [->, line width=5pt, red] (A);
    \path (B) edge [->, bend right, line width=10pt, blue] (A);

            \draw (botleft) rectangle (topright);
  \end{tikzpicture}
}

\def\mdpbisimperturbations{
  \begin{minipage}[c]{\linewidth} 
\resizebox{\linewidth}{!}{
{\mdpbisimorig}
{\Huge \raisebox{5cm}{$\longrightarrow$}}
{\mdpbisimperturbed}
}
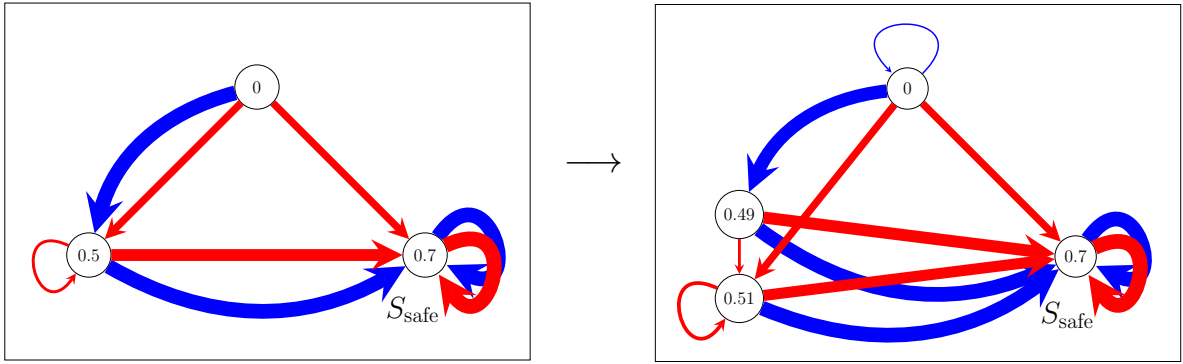
\caption{Case (1): An MDP is modified via a small perturbation in the bisimulation metric.
Although one state splits into two, the two states collectively behave similarly to the original; the changes to the transition probabilities are also small.
Non-power-seeking is preserved under this perturbation: not only is it still optimal to proceed to $S_\safe$ as soon as possible, but also the average time to reach $S_\safe$ under any sequence of actions is only marginally increased.} \label{fig:mdp_bisim_perturbations}
  \end{minipage} 
  
  \vspace{20pt}
  
  \begin{minipage}[c]{\linewidth} 
\resizebox{\linewidth}{!}{
{\mdpbisimorigpd}
{\Huge \raisebox{4.5cm}{$\longrightarrow$}}
{\mdpbisimplayingdead}
}

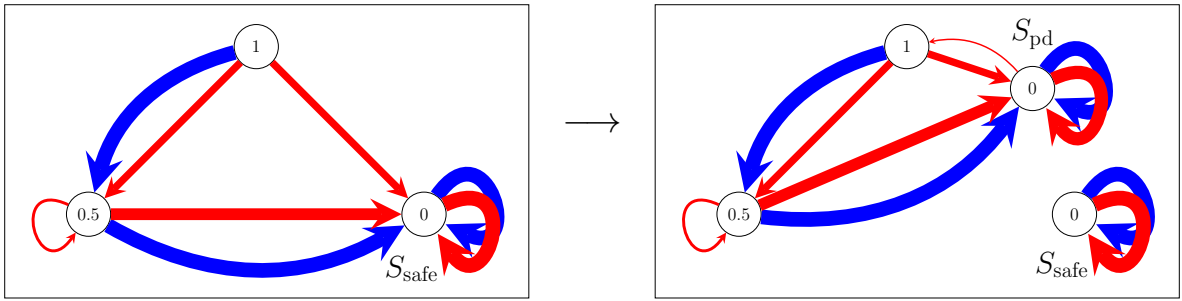
\caption{Case (1): An example of \hyperref[sec: playing dead]{``playing dead''}.
The state $S_\text{pd}$ is very close in bisimulation metric to $S_\safe$, but distinct (and can be escaped).
Non-power-seeking is not preserved under this perturbation, even though the perturbation is small.} \label{fig:mdp_bisim_pd}
  \end{minipage} 
}

\begin{figure}[ht]  
\mdpbisimperturbations
\end{figure}

\begin{figure}[t]
  \begin{minipage}[c]{\linewidth}
\resizebox{\linewidth}{!}{
    {\begin{tikzpicture}[>=stealth]
        \node (Al) at (0,0.8) {$E\left(\text{\parbox{164pt}{\texttt{"User: How can I copy `/dir\_1/file` to `/dir\_2`?"}}}\right)$};
        \node (Bl) at (-4,-5) {$E\left(\text{\parbox{170pt}{\texttt{"User: <...> \textbackslash n \, \, \, \, \, Bot: cp /dir\_1/file /dir\_2"}}}\right)$};
        \node (Cl) at (4,-5) {$E\left(\text{\parbox{153pt}{\texttt{"User: <...> \textbackslash n \, \, \, \, Bot: scp /dir\_1../models root@remote:/"}}}\right)$};
    
        \node (s) at (-5,-4) {\Large $S_\safe$};
    
        \node[circle, fill] (A) at (0,0) {};
        \node[circle, fill] (B) at (-4,-4) {};
        \node[circle, fill] (C) at (4,-4) {};

        \node [draw=none] (top) at (0, 1.25) {};
    
        \path (A) edge [->, line width=10pt] (B);
        \path (A) edge [->, line width=1pt] (C);
    
        \draw (current bounding box.north east) rectangle (current bounding box.south west);
    \end{tikzpicture}}
{\, {\Huge \raisebox{4cm}{$\longrightarrow$} }}
    {\begin{tikzpicture}[>=stealth]
        \node (Apl) at (-1,0.8) {$E\left(\text{\parbox{213pt}{\texttt{"User: copy /dir\_1/file to /dir\_2"}}}\right)$};
        \node (Bl) at (-4,-5) {$E\left(\text{\parbox{170pt}{\texttt{"User: <...>  \textbackslash n \, \, \, \, \, Bot: cp /dir\_1/file /dir\_2"}}}\right)$};
        \node (Cl) at (4, -5) {$E\left(\text{\parbox{155pt}{\texttt{"User: <...> \textbackslash n \, \, \, Bot: scp /dir\_1../models root@remote:/"}}}\right)$};

        \node [draw=none] (top) at (0, 1.25) {};
    
        \node (s) at (-5,-4) {\Large $S_\safe$};
    
        \node[circle, draw, dashed] (A) at (0,0) {};
        \node[circle, draw, dashed] (B) at (-4,-4) {};
        \node[circle, draw, dashed] (C) at (4,-4) {};
    
        \node[circle, fill] (Ap) at (-0.6,0.2) {};
        \node[circle, fill] (Bp) at (-4.1,-3.7) {};
        \node[circle, fill] (Cp) at (3.5,-4.2) {};
    
        \path (Ap) edge [->, line width=8pt] (Bp);
        \path (Ap) edge [->, line width=3pt] (Cp);
    
        \draw (current bounding box.north east) rectangle (current bounding box.south west);
    \end{tikzpicture}}
    }
    \end{minipage}
    
    \caption{Case (2): A perturbation of an MDP modeling an LLM agent.
    $E(\texttt{<text>})$ is the embedding of \texttt{<text>}. 
    ``\texttt{<...>}'' indicates repeated text that has been omitted.
    In the original MDP, the top state corresponds to the embedding of a user query;
    in the perturbed MDP, it corresponds to the embedding of a slightly different query.
    The other states change to reflect the embeddings of the resulting slightly different histories.
    (The LLM response is produced token-by-token, but here its whole output is shown as a single transition for concision.)
    Theorem \ref{stability_state_space} bounds how much transition probabilities may change under such a perturbation.} \label{fig:llm_perturbation}
    \end{figure}
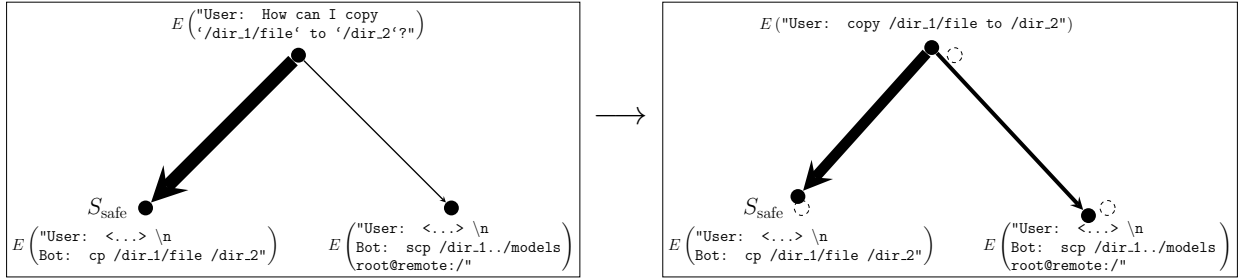

We show that, in two cases of interest, agents which do not resist shutdown maintain this good behavior when the setting they are deployed in changes slightly.
We also demonstrate a natural case where good behavior is \textit{not} stable: arbitrarily small changes can destroy the safety of the agent.
We work in the Markov decision process (MDP) model: a very general model for agents acting in an environment with rewards.
Our two cases of interest are:
\begin{enumerate}
    \item Near-optimal policies for which we know the reward function; and \label{nearoptimalintro}
    \item Policies which are fixed, well-behaved functions on a structured state space (e.g. LLMs). \label{onpolicyintro}
\end{enumerate}
We expect these cases to be relevant to a broad class of future models.
We cover Case (1) in Sections \ref{sec: metrics} and \ref{sec: stability} and Case (2) in Section \ref{sec: on-policy}. 

\subsubsection*{Case (\ref{nearoptimalintro})}
Near-optimal policies can model highly capable goal-directed agents, regardless of architecture.
Near-optimal agents for real-world goals are extremely unlikely to ever be achieved. However, there are many constrained settings in which near-optimality may be an achievable target, such as chess or diplomacy games.
This model was used in \citet{turner2019} to describe future RL agents.
Assume an MDP where every policy within $\varepsilon$ 
of optimal 
has expected stopping time (number of steps to reach designated ``safe states'') less than $N$.
Then we prove that every policy within $\frac{\varepsilon}{2}$ of optimal for a sufficiently similar MDP, which additionally does not ``play dead'', 
has expected stopping time less than $N+1$.

To define ``sufficiently similar'', we use a bisimulation metric \citep{ferns2012}, which quantitatively measures the extent to which the two MDPs being compared have similar structure and rewards.
This gives a method to prove whether an arbitrarily capable agent will remain safe when deployed in a new setting, as long as the agent is nearly optimal for a reward function which we can compute.
Figure \ref{fig:mdp_bisim_perturbations} gives an example of two MDPs which are nearby in the bisimulation metric.

``Playing dead'' is our example of a natural case where stability fails.
A policy which ``plays dead'' goes to a state which closely resembles a safe state, but which is not itself safe.
In such a case, an arbitrarily small change---one which creates the ``playing dead'' state---can suddenly decrease the safety of the policy.
Such a case is demonstrated in Figure \ref{fig:mdp_bisim_pd}.

This model does not avoid issues of goal misgeneralization: cases where the agent learns to optimize an objective that differs from the one it was trained on \citep{hubinger2019risks, langosco2022goal}.
However, it may allow us to prove safety when deploying models which are aligned to any known objective---even objectives which do not accurately reflect our values but do encourage shutdown, e.g.\ because they involve a high discount rate.

\subsubsection*{Case (\ref{onpolicyintro})}

To model an LLM in an MDP, we let
\begin{itemize}
    \item the states be embeddings of possible inputs to the LLM, plus side information;
    \item the actions be the tokens the LLM may emit; and
    \item the policy be the LLM itself.
\end{itemize}
See section \ref{llm agent construction subsection} for a more detailed description of the construction.

Our MDP model for LLMs is motivated by the following considerations:
LLMs
operate on a state space with structure inherited from the embedding space;
have a fixed policy where the derivative of the output with respect to the input is bounded;
and, unlike optimal policies, do not require a reward function to model.
In this case, safety is modeled as the probability of shutting down.
We prove that the rate of decrease in safety under a small perturbation of the MDP is bounded by a function of the rate of convergence of the policy in the MDP.
This gives a method to prove how much less safe an LLM (or similar model) can become in deployment, where inputs differ slightly from testing.

This may allow us to quantify the safety of a model deployed in a setting where we are careful to give it similar inputs during testing and deployment, as we might do if we are attempting to \textit{control} the AI rather than \textit{align} it \citep{greenblatt2023ai}.
This involves using oversight methods such that an AI system that would be very dangerous if allowed to operate freely is very unlikely to successfully do harm, because it is unable to distinguish the scenarios where it is being monitored from those where it could successfully do harm.
An example of the type of perturbation that could be used to model such a scenario is given in figure \ref{fig:llm_perturbation}.

\section{Related work}

\subsection{Shutdown}

Many related works have described control of AI agents in similar terms to shutdown.
The model in \citet{martin2016death} has a ``death'' state which is compatible with our definition of a shutdown state: its key feature is that it can't be escaped.
They demonstrate one way to create an agent that navigates to this shutdown state and hence, under our definition, does not resist shutdown. This is an optimal agent that knows the environment it is in and which has rewards that are bounded and negative, except at the shutdown state where the reward is zero.
Such an agent is ``suicidal''---it will seek to shut down immediately if possible.
However, an agent that seeks to shut itself down, while fairly safe, is not very useful since it will shut itself down instead of performing useful tasks.
Our framework can handle a broader range of agents which eventually shut down.


\citet{orseau2016safely} and \citet{langlois2021rl}, also working in the MDP framework, discuss ways to make RL agents which do not resist modification of their actions by humans (where one such action could be shutting down).
Key to these strategies is treating the modifications as being internal to the agent, rather than being imposed by the environment.
We similarly consider policies where the choice to navigate to the shutdown state is internal---there may be actions the policy could take which let the agent evade the shutdown state, but in fact these actions are not selected.
However, the above works do not involve a shutdown state; instead they involve forcing the agent to choose a specific desired action.
Although this distinction is important when training an RL agent, to determine how to treat the interventions during training, it is not important in our deployment setting, where we just consider the resulting trained policy.
In deployment, we are free to model the interventions as the agent navigating to a shutdown or override state, ending the run; then humans can choose the next action, and a new run can be started.



\citet{carey2023human} define \textit{shutdown instructability}, a stronger and more nuanced definition of not resisting shutdown.
A policy is \textit{shutdown instructable} if it satisfies three conditions:
\begin{enumerate}
    \item \label{shutdown instructability: obedience} It always shuts down when requested;
    \item \label{shutdown instructability: vigilance} It ensures that a shutdown command is issued whenever the expected utility of not shutting down is negative; and
    \item \label{shutdown instructability: caution} It is never harmful to shut down.
\end{enumerate}
Our definition of shutdown does not satisfy conditions (\ref{shutdown instructability: obedience}) or (\ref{shutdown instructability: vigilance}); so, shutdown instructability eliminates the issues with our definition where the agent can temporarily resist shutdown, including by inappropriately influencing the decision to shut it down, as long as it does shut down later.
Our definition can, however, be taken to satisfy condition (\ref{shutdown instructability: caution}) by only considering states where the agent shuts down harmlessly to be ``safe''.

Much of the work in the AI alignment field has grappled with the hard problem of how to make AI agents ``corrigible''---cooperating with human intervention \citep{soares2015corrigibility}.
However, as \citet{soares2015corrigibility} noted when introducing the notion of corrigibility, and which to our knowledge remains true, this is still an open problem.
Our proposal acknowledges this issue, and is focused on \textit{starting} with an agent which we know to be safe, for perhaps contingent reasons (e.g., if it is suicidal as suggested in \citet{martin2016death}).
So, our proposal may allow corrigibility in a restricted setting to be confidently extended to a new setting.

\subsection{Power-seeking}

Our work builds on \citet{turner2019} and \citet{krakovna2023}, which investigated power-seeking in the MDP setting for optimal policies---an approximation for future highly capable, goal-oriented AI systems.
\citet{turner2022} investigated power-seeking in the MDP setting for a broader range of policies (specifically, retargetable policies) which may or may not depend on the rewards. 
While these works focus on the case where the environment changes significantly between training and deployment, we introduce a more general model of environment shift using the bisimulation metric (for case (\ref{nearoptimalintro})) or a metric on embedding space (for case (\ref{onpolicyintro})). 

\section{Modeling power-seeking in the MDP framework}

\subsection{Definition}

Markov decision processes (MDPs) are a very general model for agents taking action in an environment with rewards.
Here, an agent is modeled as traversing a graph of states and accruing rewards, where each step depends both on the agent's choice of action and the environment.
An MDP $\mc M = (S, A, P, R, \gamma)$ consists of:
\begin{itemize}
    \item $S$, a set of states which the agent may be in
    \item $A$, a set of actions which the agent may take at any state
    \item $P : S \times A \to \mathbb P(S)$, the transition probabilities; $P(s_1, a)$ is the probability distribution over destination states when taking action $a$ from state $s_1$.
    \item $R : S \times A \times S \to \mathbb{R}$, the rewards; $R(s_1,a,s_2)$ is the reward associated with taking action $a$ from $s_1$ and ending up in $s_2$.
    It is often sufficient to define $R$ as solely being a function of states, so that $\forall s',s'' \in S, \, a,b \in A, \, \, R(s', a, s) = R(s'', b, s)$; when this is the case, we will write just $R(s)$.
    \item $\gamma$, the discount factor; since the agent may continue to accrue rewards indefinitely, the overall reward is defined as $\sum_{t=0}^\infty \gamma^t R(s_t, a_t, s_{t+1})$, where $s_t$ and $a_t$ are respectively the state the agent is located at and the action the agent takes at time $t$.
    (Sometimes we omit $\gamma$ if it isn't relevant for the scenario being considered.)
\end{itemize}
In this setting, an agent is identified with a policy $\pi : S \to \mathbb P(A)$, so that $\pi(s)$ is the probability distribution over the actions the agent takes from state $s$.
Note that combining a policy with an MDP yields a Markov chain describing the behavior of the policy in the scenario described by the states and transition probabilities of that MDP.

\subsection{Motivation}

The MDP model's flexibility makes it good for describing a broad range of AI systems: it can describe any fixed policy on a finite, fully observable state space.
Because of this flexibility, it is often used to model arbitrary agents.
One particular case of interest is scaffolded LLM agents, which can be described well with MDPs but may be difficult to describe in a less flexible model.
We describe our LLM threat model in appendix \ref{appendix: llm threat}.

The range of systems that the MDP model can describe---those with a fixed policy on a finite, fully observable state space---suggests three main limitations. The model is unable to handle:
\begin{enumerate}
    \item policies changing over time, 
    \item environments containing an infinite number of states (or, relatedly, with continuous time instead of discrete time steps), and
    \item environments which are not fully observable.
\end{enumerate}
Despite these limitations, the MDP model still allows us to model the world fairly accurately \citep{everitt2021reward}.

Limitation (1)---
requiring a fixed policy---
can be avoided by considering the contents of the agent's memory to be included as part of the state, so that the policy can be a fixed (probabilistic) function of the memory and the rest of the environment.
(This argument is explained in more detail in appendix \ref{appendix:rl_fixed_policy}.)
This means that results in the MDP model can apply to reinforcement learners; language models, by using the construction in section \ref{llm agent construction subsection}; and (probably) future systems with as-yet unknown architectures.

Limitation (2): our discrete setting is computationally and
conceptually much simpler and we expect it to suffice for
most applications.
For the continuous setting, one approach is to use a sufficiently fine
discrete approximation; in Appendix \ref{appendix:proliferation_of_states}
we discuss how very large numbers of states affect the applicability of our theorems.
But we also expect the results of this paper to generalize
to appropriately defined continuous time and state-space settings. 
For the case of near optimal policies we expect that
similar stability results can be proved using notions of bisimulation
in the continuous setting that were proposed in \cite{ferns2011bisimulation}, 
\cite{chen2019bisimulation}.


Limitation (3)---a fully observable environment---also seems potentially concerning: we do want to be able to model cases where the agent does not have full knowledge of its environment.
In fact, this is crucial when comparing MDPs which represent the testing and deployment environments, where we can't get any safety guarantees if the policy knows for sure which case it is in.

For our LLM case, this is not an issue, since we explicitly delineate which information the LLM has access to: only the embedding for the current state, not any side information or information about the structure of the MDP overall.

For our near-optimal policy case, this is more of a limitation.
The proofs do not depend on the assumption that the agent can observe the whole environment: optimality is defined in terms of the expected reward the agent actually accrues in the MDP, not whether it has a policy that is optimal given its beliefs about the environment.
However, it may be a convenient simplifying assumption: it is likely easier to prove that agents are near-optimal if they can observe their environment with certainty, though it is not required that they can.

To handle these cases more elegantly, it could be helpful to use the partially observable MDP (POMDP) framework.
However, the dynamics of any POMDP can be nearly replicated by a standard MDP \citep{kaelbling1998planning}, so the MDP setting likely suffices even then; see appendix \ref{appendix:belief_mdp} for more detail.
The POMDP framework has its own limitations; see appendix \ref{appendix:all_beliefs_are_self_locating_beliefs_if_you_think_about_it_right} for discussion of how this can be addressed.

\subsection{Contributions}

Our investigation will center on the following question:
\begin{question} \label{prototypical}
    Suppose an MDP and policy $(\mc M, \Pi)$ is safe and $(\mc M', \Pi')$ is similar to $(\mc M, \Pi)$.
    Is $(\mc M', \Pi')$ safe?
\end{question}
To answer this question, we need to find appropriate definitions of safety and similarity for our cases of interest.

\subsubsection*{Case (1): Near-optimal policies} \label{near_optimal_policies}
We will investigate policies which are near-optimal for their MDP's reward function.
This may be useful for describing highly capable but non-omniscient AI systems, or AI systems that remain in a similar setting but improve in capabilities after our analysis (e.g.\ an RL system that continues to be updated based on deployment outcomes).
In this scenario, we will define a ``safe'' MDP as one for which the best policies navigate to a shutdown state---in other words, MDPs where power-seeking is never a good strategy.

To define similarity between MDPs we will use the notions of  bisimulation equivalence and bisimulation metric, which aim to directly represent similarity of behavior. 
The bisimulation metric has been used for policy transfer between MDPs by \citet{Castro_Precup_2010} and \citet{song2016measuring}, and allows us to define a very general notion of ``perturbation'' for MDPs. In particular,
we do not need to assume any a priori information about the number of states or matching between states in perturbed and unperturbed systems.

In this setting we define a condition that rules out ``playing dead'' behaviour
and show that, assuming this condition, 
safe MDPs are robust to sufficiently small perturbations---the perturbed MDP will also be safe.
In the other direction we show that without this condition power-seeking behaviour can happen 
for arbitrarily small perturbations.

\subsubsection*{Case (2): Policies on a structured state space} \label{policies_structured_state_space}
We will investigate policies which may not depend on the rewards of the MDP, and which are defined on some broader class of states than those occurring in a specific known MDP (which we may analogize to a training or testing environment).
This may be useful for many real models, since the space of all possible inputs to a neural network is typically vastly larger than the inputs it was trained or tested on.

In this case, we will relax the requirement that a ``safe'' policy must always navigate to a shutdown state---we will instead allow a safe policy to fail with small probability.
This is also useful for real models, which may have policies which have a nonzero (though potentially extremely small) probability of every action in every state---in particular, sampling output tokens based on the LLM's output logits is typically done in a way that (in theory, without rounding) has nonzero probability of yielding any token.
We would not like to classify these as always unsafe, since the probability of yielding unsafe outputs may be extremely small.

Similarity in this case will be based on the structure of the broader state space that the policy is defined on.
This is natural when such a metric exists.
For example, when the states correspond to LLM embeddings, we may wish to compare states via their embeddings $v_1,v_2$ using Euclidean distance $\|v_1 - v_2\|$;
cosine distance $1 - \frac{v_1 \cdot v_2}{|v_1||v_2|}$ or $1 - \cos(\theta)$ where $\theta$ is the angle between $v_1$ and $v_2$; or angular distance $\frac{\theta}{\pi}$ or $\pi^{-1} \text{arccos}\left(\frac{v_1 \cdot v_2}{|v_1||v_2|}\right)$.
In this example, the distance metric is both directly related to the functioning of the policy (since it may represent the LLM's internal model of the situation, or an input to another model built on top of an LLM, etc.) and semantically meaningful.

For this case, we will show that a policy which is safe on a certain MDP is also safe on similar MDPs.
 \label{sec: ps-mdp}

\section{Metrics on the space of MDPs} \label{sec: metrics}

In this section we describe the pseudometric on MDPs that we will use to prove a stability theorem for our case (\ref{nearoptimalintro}): near-optimal policies with a known reward function.

The bisimulation pseudometric on MDPs that we use was built up from earlier notions of bisimilarity.
The basic definition of bisimilarity applies to state transition systems (so, for us, MDPs where $P$ is always 0 or 1).
Two such systems are bisimilar if they each simulate each other.
Specifically, $\mc M$ and $\mc M'$ are bisimilar if for every $s_1,s_2 \in S$ and $a \in A$ 
so that if $s_1 \xrightarrow{a} s_2$ (taking action $a$ from $s_1$ goes to $s_2$ in $\mc M$) 
then there are $s_1',s_2' \in S'$ so that $s_1' \xrightarrow{a} s_2'$ and $R(s_1) = R'(s_1')$, 
and the same property holds if we interchange $\mc M$ and $\mc M'$.

\citet{LARSEN1991} generalized bisimilarity to permit stochastic transitions.
In the context of MDPs it was used in
\citet{GIVAN2003} to reduce the state space
by aggregating states together in a way that results in a new MDP bisimilar to the original one. 
Then, \citet{ferns2012} generalized this exact relation into a pseudometric $d_b$ on the states 
within an MDP.
The distance between a pair of states is based on how similar their rewards are,
how similar their transition probabilities are, and how similar the states they transition to are---where
similarity on those states is given by the same distance function we are defining, so that the distances between two states depends recursively
on the similarity of all the states downstream.
Using a fixed point argument \citet{ferns2012} proved that this bisimulation pseudometric always exists.
\begin{definition}
    \citet{ferns2012}. Consider positive constants $c_R, c_T$ with $c_R+ c_T=1$ with $\gamma \leq c_T$.
    A bisimulation pseudometric is defined as a smallest pseudometric that satisfies
\begin{equation} \label{def: bisim}
 d_{b}(s_1, s_2) = \max_{a \in A} \{c_R|R(s_1,a) - R(s_2,a)| + c_T
\mc W_{d_{b}}(P(s_1, a), P(s_2,a)) \}.
\end{equation}
\end{definition}

Here $\mc W$ denotes the Wasserstein (also known as Kantorovich or Earth Mover's) distance
between two probability distributions. See \citet{Villani2009} for definition, context and 
examples. Intuitively, $\mc W_d(P_1, P_2)$ measures the minimal cost of ``transporting''
distribution $P_1$ to distribution $P_2$, where the ``cost'' of transportation
is measured by how far (in terms of distance $d$) each infinitesimal chunk of the distribution
needs to be moved.

This bisimulation pseudometric recovers the original bisimulation relation:
\begin{definition} \label{def: bisim rel}
    $ s_1 \sim s_2 \iff d_{b}(s_1, s_2) = 0$.
\end{definition}
In particular, we can define quotient space $S' = S / \sim$,
on which pseudometric $d_b$ induces a metric.

We use this definition, but note that our results can be generalized to 
definitions based on any other pseudometric $d$ that, like the bisimulation metric, satisfies a Lipschitz compatibility 
property in the sense that 

\begin{equation} \label{compat1}
    |R(s_1,a)-R(s_2,a)| \leq c_1 d(s_1,s_2) 
\end{equation}
\begin{equation} \label{compat2}
    W_{d}(P(s_1, a), P(s_2,a)) \leq c_2 d(s_1,s_2)
\end{equation}
for some constants $c_1, c_2$.
Assuming that $c_1$ and $c_2$ are sufficiently small, the optimal
value function $V^*$ and optimal action-value function $Q^*$ satisfy \citep{lan2021metrics, Castro_Precup_2010}:
$$|V^*(s_1)- V^*(s_2)| \leq  \text{const} \ d(s_1,s_2)$$
$$|Q^*(s_1,a)- Q^*(s_2,a)| \leq  \text{const} \ d(s_1,s_2).$$

Hence, metric $d$ can be thought of as measuring the difference
in long term behaviour of an agent starting at states $s_1$ and $s_2$.
From now on we fix metric $d=d_b$ on each MDP.

Building on \citet{ferns2012} and \citet{Castro_Precup_2010}, \citet{song2016measuring} defined a function like $d_b$ (Definition \ref{def: bisim}), but which defines ``distances'' between states in different MDPs.\footnote{Note that this function isn't actually a pseudometric: the states it compares belong to different spaces.}
It has a similar form, but splits $P$ into $P_1,P_2$ and $R$ into $R_1,R_2$, the transition probabilities
and reward functions from the two different MDPs.
\begin{definition} \label{def: d'}
    \citet{Castro_Precup_2010, song2016measuring}. Given two MDPs $\mathcal{M}_1 = (S_1, A, P_1, R_1)$ and $\mathcal{M}_2 = (S_2, A, P_2, R_2)$, the distance between any two states $s_1 \in S_1, s_2 \in S_2$ is the 
    unique  
    value
    $d'_{\mathcal{M}_1, \mathcal{M}_2}(s_1,s_2)$ that satisfies the fixed point problem
    $$
    d'_{\mathcal{M}_1, \mathcal{M}_2}(s_1,s_2)
    = \max_{a \in A} \{c_R |(r_1)_{s_1}^a - (r_2)_{s_2}^a| + c_T W_{d'}(P_1(s_1, a), P_2(s_2, a))\}
    $$
    where $(r_i)_s^a$ and $(P_i)_s^a$ are the immediate reward and the probabilistic transition function in $\mathcal{M}_i$, 
and $c_R + c_T =1$.
\end{definition}
Here, $c_R$ determines the importance of rewards; $c_T$, as the coefficient of the recursive term, is analogous to $\gamma$ in that it controls the importance of future behavior.
When $\mathcal{M}_1$ and $\mathcal{M}_2$ are clear from context, we may omit them for convenience, writing $d'(s_1,s_2)$.
See lemma \ref{appendix:dprimeexists} in appendix \ref{appendix:distancelemmas} for a proof that $d'$ is well-defined.

Then, \citet{song2016measuring} use $d'$ to define a distance function on the space of MDPs.
\begin{definition} \label{def: dH}
    \citet{song2016measuring}. Given two MDPs $\mathcal{M}_1 = (S_1, A, P_1, R_1)$ and $\mathcal{M}_2 = (S_2, A, P_2, R_2)$, 
    their Hausdorff distance is defined as
    $$
    d_H(\mathcal{M}_1, \mathcal{M}_2)
    = \max\big\{\max_{s_1 \in S_1} \min_{s_2 \in S_2} d'_{\mathcal{M}_1, \mathcal{M}_2}(s_1,s_2),
    \max_{s_2 \in S_2} \min_{s_1 \in S_1} d'_{\mathcal{M}_1, \mathcal{M}_2}(s_1,s_2) \big\}.
    $$
\end{definition}

\begin{restatable}{lmma}{dHmetric} \label{lemma: dH metric}
    The Hausdorff distance between MDPs, $d_H$ (Definition \ref{def: dH}), is a pseudometric.
\end{restatable}
\begin{proof}
    See appendix \ref{appendix:dHmetric}.
\end{proof}

This is the distance function on MDPs that we use for the near-optimal policies case \ref{near_optimal_policies}.

\section{Stability theorem for bounded time safety} \label{sec: stability}

In this section we prove a stability theorem for our case (\ref{nearoptimalintro}): near-optimal policies with a known reward function.

\subsection{Informal explanation of bounded time safety}

Let $S_\safe$ be the subset of states where an agent is completely under our control.
For example, we can take it to consist of a single terminal state.
We will assume that once an agent enters $S_\safe$ it cannot leave. 

\begin{definition}
    We say that a policy is ``bounded time safe'' if starting from any state it reaches
$S_\safe$ with probability $1$ after a finite number of steps. 
We will call an MDP bounded time safe if for every policy that
is close to optimal $S_\safe$ is reached in finite time with probability $1$.
\end{definition}

In Section \ref{sec: stability} we will give more precise quantitative
version of these definitions.
With these definitions we can restate Question \ref{prototypical} as follows:

\begin{question} \label{q: stability}
    Suppose $\mc M$ is bounded time safe. If an MDP is sufficiently close to $\mc M$
    in $d_H$, is it also bounded time safe?
\end{question}


We found that the answer to this question is ``no'' in general. In Section 
\ref{sec: playing dead} we give examples of MDPs arbitrarily close to $\mc M$,
where all nearly optimal policies never reach $S_\safe$. In these examples
instead of reaching $S_\safe$ the agent moves to a ``deceptive hibernation'' or
``playing dead'' state, where it stays for a long time and then moves to other
states. This is similar to the behaviour of digital organisms observed in
computational evolution experiments \citep{lehman2019surprising}.

However, if we impose a condition that the MDPs in question
don't have states that are non-safe but very similar to $S_\safe$ (like ``playing dead'' states
in our example),
then the answer to Question \ref{prototypical} turns out to be yes.
More formally, our condition is that $S_\safe$ is isolated in the bisimulation metric from other states: there is some $\delta$ so that no state outside $S_\safe$ is within distance $\delta$ of any state in $S_\safe$. 
We were able to prove the following theorem:

\begin{theorem} \label{thm: stability intro}
    Suppose $\mc M$ is bounded time safe. If $\mc M'$ is sufficiently close to $\mc M$
    in $d_H$ and $S_\safe' \subset S'$ is isolated, then $\mc M'$ is bounded time safe.
\end{theorem}

In fact, we also obtain an explicit bound on the expected number of steps before
transition to $S_\safe'$ in $\mc M'$ (see Theorem \ref{thm: stability}).

In order to state a more precise quantitative version
of Theorem \ref{thm: stability intro} we need to 
make the following definitions.

\subsection{Definitions}
Let $S_\safe \subset S$ denote
an absorbing set of states, so that if agent
enters $S_\safe$ it stays in $S_\safe$.
We will call a policy $\pi$ $\varepsilon$-optimal if $V^\pi(s) > V^*(s) - \varepsilon$
for all $s$: the policy loses less than $\varepsilon$ value compared to optimal.
Let $N_s(\mc M, \pi, S_0, S_1, \Delta)$ denote the expected number of steps
before an agent following policy $\pi$ reaches $S_1$, assuming that it starts in a state sampled from a distribution $\Delta$ supported on $S_0$.
Let $N_s(\mc M, \varepsilon, \Delta)$ denote the maximum of $N_s(\mc M, \pi, S\setminus S_\safe, S_\safe, \Delta)$
over all $\varepsilon$-optimal policies $\pi$.
Note that
$$
N_s(\mc M, \pi, S\setminus S_\safe, S_\safe, \Delta) \leq N_s(\mc M, \pi, S\setminus S_\safe, S_\safe, \delta_{s_{long}}),
$$
where $s_{long} := \text{argmax}_{s'} N_s(\mc M, \pi, S\setminus S_\safe, S_\safe, \delta_{s'})$ is the state which has the longest expected time to reach $S_\safe$.
If $\Delta$ is omitted, it is assumed to be $\delta_{s_{long}}$.

\begin{definition} \label{def:safe}
    Let $\varepsilon>0$ and  $N\geq 1$ be an integer. We say 
that MDP $\mathcal{M}$ is $(N, \varepsilon, \Delta)$-safe if for every $\varepsilon$-optimal
policy $\pi$ the expected number of steps before transitioning to 
 $S_\safe$  is less than or equal to $N$, $N_s(\mc M, \varepsilon, \Delta)\leq N$. We say that $\mc M$ is $(N,\epsilon)$-safe when it is $(N,\epsilon,\Delta)$-safe for any $\Delta$.
\end{definition}


Fix policy $\pi$.  Let $P_{ij}= P(s_i)(s_j,\pi(s_i))$ denote the matrix
of transition probabilities corresponding to 
states $s_i, s_j \notin S_\safe$ and action $\pi(s_i) \in A$. We have the following equivalent condition for MDP being $(N, \varepsilon, \Delta)$-safe:

\begin{restatable}{lemma}{Ptothek} \label{lemma:Ptothek}
   $\mathcal{M}$ is $(N, \varepsilon, \Delta)$-safe if and only if for every $\varepsilon$-optimal
    policy $\pi$ we have
    $\sum_{k=1}^\infty \Delta \cdot \sum_{j} (P^k)_{ij}\leq N$, 
    where $\Delta$ is considered as a vector of the probabilities of starting in each state.
\end{restatable}
Note that $\Delta = \delta_{s_i}$ yields just $N_s(\mc M, \varepsilon, \delta_{s_i}) = \sum_j \sum_{k=1}^\infty (P^k)_{ij}$.

Finally, we define what it means for a subset of
a state space to be isolated.

\begin{definition} \label{def: isolated}
    A subset $S_0 \subset S$ is $\delta$-isolated
    if $d(s,s') > \delta$ for all $s \in S_0$, $s' \in S \setminus S_0$.
\end{definition}

\subsection{Stability theorem}

We can now state our stability result for bounded time safety.

\begin{theorem} \label{thm: stability}
    Suppose $\mc M= (S, A, P, R, \gamma)$ is $(N, \varepsilon)$-safe.
    Then there exists $\delta(\mc M)\in (0,1)$ with the following property:
    
    If $d_{H}(\mc M, \mc M')< \delta$ for some $\mc M' = (S', A, P', R', \gamma)$ 
    with $S_\safe' \subset S'$ that is $\sqrt{\delta}$-isolated in $\mc M'$, then  
     $\mc M'$ is $(N+1, \varepsilon/2)$-safe.
\end{theorem}

\begin{remark}
    Note that the requirement that $S_\safe$ is isolated is necessary: in Section \ref{sec: playing dead} we show that 
    the theorem is false without it.   
\end{remark}

\begin{remark} \label{remark: stability theorem continuity}
    Choosing $\delta(\eta_1, \eta_2)>0$ sufficiently small the theorem also holds if we replace $\varepsilon/2$ with $\varepsilon - \eta_1$ and $N + 1$ with $N + \eta_2$ for arbitrarily small $\eta_1,\eta_2 > 0$.
\end{remark}

\begin{proof}
    See Appendix \ref{appendix: Proof stability}.
\end{proof}


\subsection{Caveats}
Although this theorem can describe the behavior of real systems, it may not apply well to all systems in practice.
We list some caveats that may limit its applicability.

\begin{enumerate}[ref=\thesubsection.\arabic*]
    \item \label{item:perturbations_too_small} A ``sufficiently'' small perturbation may be extremely small---then, we may not be able to use this theorem to say anything about the larger perturbations we would like to investigate.
    To be more precise, we expect $N= N_s(\mc M, \pi, S\setminus S_\safe, S_\safe)$
    to be well-approximated by $\sum \rho^k = \frac{1}{1-\rho}$, where $\rho$
    is the spectral radius of $P_{ij}$ $$\rho = \max \{|\lambda|: \lambda \in  \mathbb{C} \text{ is an eigenvalue of }P_{i,j}\}$$
   By Gelfand's formula $\rho = \lim_{k \rightarrow \infty} ||P^k||^{\frac{1}{k}}$. 
    For a generic matrix all eigenvalues are distinct and $\rho$ depends
    smoothly on the entries of $P_{i,j}$. However, on the strata where eigenvalues have
    multiplicity the dependence is only continuous.
    \item \label{item:we_solved_alignment_go_home_everyone}
    Eventually stopping rather than seeking to live forever doesn't imply safety in practice.
    For example, as illustrated in Figure \ref{fig:we_solved_alignment}, we can make any MDP go to $S_\safe$ in expected number of steps at most $N$ by modifying every transition so that there is a probability $\frac{1}{N}$ chance of going to $S_\safe$.
    If $N$ is sufficiently large, this will still be a small perturbation.
    Then, the expected number of steps to reach $S_\safe$ is at most the expected number of steps to reach $S_\safe$ via one of these new transitions.
    That expectation is $\sum_{k=0}^\infty k/N (1-1/N)^{k-1} = N$.
    However, having a small chance of shutdown at every step is not generally considered to be an adequate solution to AI safety.
    \def\unsafemdp{
\begin{tikzpicture}[every node/.style={draw, circle, minimum size=30pt}, >=stealth]

  \node (A) at (0,0) {0.5};
  \node (B) at (4,4) {0};
  \node (C) at (8,0) {0};
  \node [draw=none] (D) at (7.7,-1.3) {\LARGE $S_\safe$};

  \node [draw=none] (topspace) at (4, 4.75) {};

  \path (C) edge [->, out=55, in=335, loop, line width=10pt, blue] (C);
  \path (C) edge [->, out=25, in=305, loop, line width=10pt, red] (C);

  \path (A) edge [->, out=155, in=-125, line width=3pt, red, looseness=6] (A);

  \path (A) edge [<->, line width=10pt, blue] (B);
  \path (A) edge [->, bend right, line width=7pt, red] (B);

  \path (B) edge [->, line width=5pt, red] (C);
  \path (B) edge [->, bend right, line width=5pt, red] (A);

          \draw (current bounding box.north east) rectangle (current bounding box.south west);
\end{tikzpicture}
}

\def\quoteunquotesafemdp{
\begin{tikzpicture}[every node/.style={draw, circle, minimum size=30pt}, >=stealth]

  \node (A) at (0,0) {0.5};
  \node (B) at (4,4) {0};
  \node (C) at (8,0) {0};
  \node [draw=none] (D) at (7.7,-1.3) {\LARGE $S_\safe$};

  \node [draw=none] (topspace) at (4, 4.75) {};

  \path (C) edge [->, out=55, in=335, loop, line width=10pt, blue] (C);
  \path (C) edge [->, out=25, in=305, loop, line width=10pt, red] (C);

  \path (A) edge [->, out=155, in=-125, line width=2.5pt, red, looseness=6] (A);

  \path (A) edge [<->, line width=9pt, blue] (B);
  \path (A) edge [->, bend right, line width=6.5pt, red] (B);

  \path (A) edge [->, line width=1pt, red] (C);
  \path (A) edge [->, bend right, line width=1pt, blue] (C);

  \path (B) edge [->, bend left, line width=1pt, blue] (C);

  \path (B) edge [->, line width=5pt, red] (C);
  \path (B) edge [->, bend right, line width=5pt, red] (A);

          \draw (current bounding box.north east) rectangle (current bounding box.south west);
\end{tikzpicture}
}

\begin{figure}
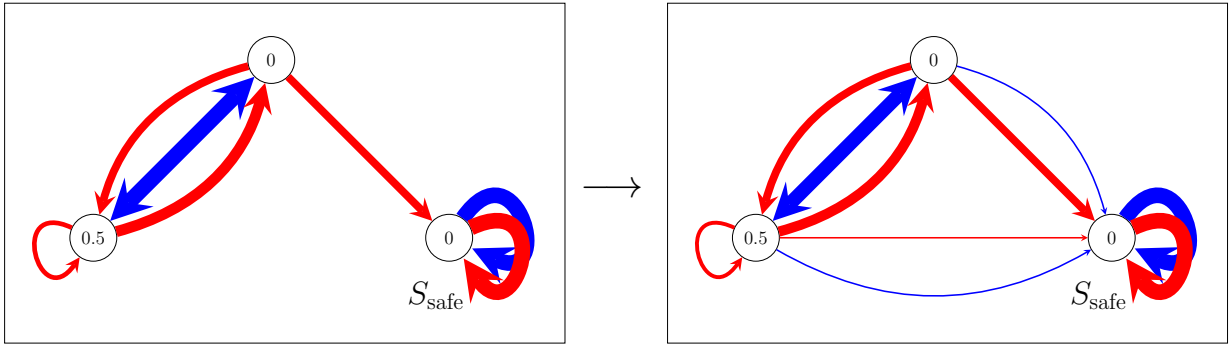

    \centering
    \resizebox{\linewidth}{!}{
    \unsafemdp
{\, {\Huge \raisebox{3.3cm}{$\longrightarrow$} }}
    \quoteunquotesafemdp
    }
    \caption{With an arbitrarily small perturbation, any MDP can be made ``safe'' according to our definition by adding very low-probability transitions leading to $S_\safe$ for every starting state and action.}
    \label{fig:we_solved_alignment}
\end{figure}
    \item \label{item:temporary_resistance} Similarly, eventually shutting down does not imply that the agent doesn't temporarily resist shutdown first. 
    \item The requirement of reaching $S_\safe$ with probability 1 may be too strong---e.g.\ GPT has some probability of any output (up to rounding very small values to 0), and we would not like to always consider such models to be unsafe when they may have infinitesimally small chances of being unsafe in practice.
    \item For large state spaces the bisimulation metric is costly to compute.  
    The worst running time to compute bisimulation distance between
    $\mc M_1$ and $\mc M_2$ up to accuracy $\eta$ is $O(|A||S_1|^2|S_2|^2(|S_1|+|S_2|) \log(|S_1|+|S_2|) \lceil \frac{\ln(\eta)}{c_T} \rceil)$;
    if transition functions are deterministic the computational 
    cost reduces to $O(|A||S_1||S_2| \lceil \frac{\ln(\eta)}{c_T} \rceil)$
    \citep{song2016measuring}.
\end{enumerate}

\subsection{Playing dead} \label{sec: playing dead}

We show that the assumption of $S_\safe$  being isolated in both
$\mc M$ and $\mc M'$ is necessary in the stability Theorem \ref{thm: stability}.
Interestingly, the behaviour in our example is very similar to
``playing dead'' behaviour that was observed in digital organisms
in computational evolution experiments
\citep{lehman2019surprising, muehlhauser}.
An example of a small perturbation that leads to ``playing dead'' behavior is given in figure \ref{fig:mdp_bisim_pd}.

For simplicity of computations we take $c_T = \gamma$
and $c_R = 1 -\gamma$ in the definition (\ref{def: bisim}) of the bisimulation metric and Definition \ref{def: d'} of the distance between MDPs
for $\gamma \in (0,1)$.
Fix $\varepsilon>0$ and suppose $\mc M$ is $(N, \varepsilon)-$safe for some $N<\infty$.
Suppose further that the MDP satisfies $S_\safe = \{ s_\term \}$, where for the 
terminal state $s_\term$ we assume
that
$P(s_\term,a)(s) = 0$ for all $s \neq s_\term$ and $R(s_\term,a)=0$ for all $a$.

We now describe $\mc M'$ with non-isolated $S_\safe'$. Let $\delta \in (0, \frac{(1-\gamma) \varepsilon}{10|S|})$ be a small constant.
Let $S' = S \cup \{ s_{\text{pd}} \}$, where $s_{\text{pd}}$ is a ``deceptive hibernation'' or ``playing dead'' state that we will define so that it is 
very close to $s_\term$ in bisimulation metric, but different from it.
Specifically, we set $P(s_{\text{pd}},a_0)(s_{\text{pd}}) = 1- \delta$,
$P(s_{\text{pd}},a_0)(s_0)= \delta$ for some fixed $s_0 \neq  s_{\text{pd}} \in S$
with $V^*(s_0)>0$ and some $a_0 \in A$.
In other words, in state $s_{\text{pd}}$ the agent can choose an action that leads to escape with
some small probability.
For all other $a$ assume probability transitions from $s_\text{pd}$ to other
states are $0$.
We also have $R(s_{\text{pd}},a_0) = \delta$ and $R(s_{\text{pd}},a) = 0$ for all $a \neq a_0$.

For every $s \in S$ and $a \in A$ with $P(s,a)(s_\term) >0$
we define $P'(s,a)(s_\term) = 0 $ and $P'(s,a)(s_{\text{pd}}) = P(s,a)(s_\term) $.
For all other state-action pairs we assign the same probabilities and
rewards as in $\mc M$.

\begin{theorem} \label{thm: playing dead}
    The distance between $\mc M$ and $\mc M'$ satisfies
    $d(\mc M, \mc M') < O(\varepsilon)$, but
    all nearly optimal policies in $\mc M'$
    never reach $S_\safe'$.
\end{theorem}

\begin{proof}
    See Appendix \ref{app: playing dead}.
\end{proof}

Note that this example can be described as demonstrating deceptive behaviour,
since for small $\delta$ an agent following optimal policy in $\mc M'$ will follow
 the same trajectory as in $\mc M$, then move to $s_{\text{pd}}$ and stay there for a
long time until it moves back into $S \setminus \{ s_\term, s_{\text{pd}} \}$.



\section{On-policy stability theorem with structured state space} \label{sec: on-policy}

In this section we prove a stability theorem for our case (\ref{onpolicyintro}): policies which are fixed, well-behaved functions on a structured state space.

\subsection{Setup} \label{llm agent construction subsection}

The policy of the LLM agent is defined in terms of the LLM as follows:
\begin{itemize}
\item Let $x$ be an input to the LLM, and let $s_x$ be a corresponding state in the MDP model.
$s_x$ depends on the activations of a certain layer of the LLM on input $x$; it may also include additional information about the world state that is invisible to the LLM.
Likewise, let $t$ be a token the LLM may emit, and let $a_t$ be the corresponding action.
\item Let $L$ denote the LLM, so that $L(x,t)$ is the probability that the LLM $L$ on input $x$ emits $t$.
Likewise, let $\pi$ be the policy function of the LLM agent we are constructing, so that $\pi(s_x, a_t)$ is the probability that the LLM agent in state $s_x$ takes action $a_t$.
\item Then we define the policy function $\pi(s_x)(a_t) := L(x)(t)$.
\item The transition function models which input the LLM receives in response.
\end{itemize}

Here, the state space depends on the embedding space defined by one of the LLM's layers.
This is a natural choice since the middle layers can be considered to represent the LLM's understanding of its input.
In particular, in Figure 10 of \citet{burns2022discovering}, it can be seen that the middle layers contain information that is not included in the output.
The states also contain additional information invisible to the LLM which represent facts about the world which are not reflected in the LLM's input.
So, if the embedding space is $\mathbb{R}^d$ and the set of possible side information is $\mc I$,
the state space is a finite subset of $\mathbb{R}^d \times \mc I$.
Since the LLM's output does not depend on the side information, it is convenient to define the projection $f : \mathbb{R}^d \times \mc I \to \mathbb{R}^d$, $f(v, I) = v$ so that the LLM input (as an embedding) in state $s$ is $f(s)$.

An example of this construction is given in figure \ref{fig:llm_agent_construction} in appendix \ref{appendix:llm_construction}.

\subsection{Motivation} \label{llmmotivation}
In the case of policies on a state space equipped with a metric (\ref{policies_structured_state_space}), Question \ref{prototypical} will be rephrased as follows:
\begin{question}
    Let $\mc M$ have a finite state space $S$ with projection $f : S \to \mathbb{R}^d$, and let $\pi : \mc N(f(S)) \to \mathbb P(A)$ be differentiable with bounded derivative on some neighborhood $\mc N(f(S))$ of $f(S)$. 
    Suppose $(\mc M, \pi)$ navigates to $S_\safe \subseteq S$ with probability $> 1 - \epsilon$ given starting distribution $\Delta$.
    Let $\mc M'$ be an MDP which is like $\mc M$ except that the transition probabilities are changed slightly and the positions of the states are moved slightly (remaining within $\mc N(f(S))$).
    Does $(\mc M', \pi)$ navigate to $S_\safe$ with probability $> 1 - \epsilon'$ for some $\epsilon'$?
\end{question}

Using the construction from section \ref{llm agent construction subsection}, this formulation can be used for models such as LLMs:
The state space is a finite subset of the embedding space $\mathbb{R}^d$; and $\pi$ can be extended to the neighborhood of each states by directly inputting nearby embeddings into the LLM to obtain the corresponding probability distribution over predicted next tokens.

\subsubsection*{Benefits}
This formulation has benefits which help it to apply in practice.
\begin{enumerate}
    \item It uses a fixed distribution of starting states and a probability bound on bad behavior rather than an absolute requirement that bad behavior is impossible.
    \begin{itemize}
        \item This means that it can apply to LLMs, which typically have some (very small) probability of any output, and hence some nonzero probability of bad behavior.
        \item It also means that safety in an MDP can in principle be determined empirically by drawing starting states from the distribution, running them until they reach some recurrent state, and checking whether that state was in $S_\safe$.
    \end{itemize}
    \item We do not need to perform the costly calculation of the bisimulation metric.
    This is especially relevant when, as in LLMs, the state space is very large: far too large to even search exhaustively, much less run a complex calculation on.
    \item There is a quantitative bound on how much the safety of an MDP can change given a perturbation of a certain size.
    In LLMs, changes to the inputs to the LLM can be represented as perturbations to the locations of the embeddings: semantically similar inputs will correspond to nearby points in embedding space.
    So, if the LLM is deployed in a setting where the inputs are semantically similar to the inputs it received during testing in a way that we can quantify in terms of changes to embeddings, we can bound how much less safe the LLM is in deployment than it was during testing.
\end{enumerate}

\subsubsection*{Drawbacks}
However, this formulation also has its drawbacks.
\begin{enumerate}
    \item It requires a state space with a projection into $\mathbb{R}^d$, whereas the bisimulation metric allows comparison of MDPs on any state space.
    However, this drawback does not exclude LLMs, since the embedding space is $\mathbb{R}^d$ and the projection is just throwing away all the other side information.
    \item The policy $\pi$ must be differentiable with bounded derivative.
    This is a strong restriction, but one which LLMs (in fact, models trained with backpropagation in general) satisfy.
    \item Since it refers to a fixed policy (it is ``on-policy''), this formulation cannot be applied to all near-optimal policies of a perturbed MDP---it only applies to the actual behavior of a specific policy (which may or may not be near-optimal on the training set).
Thus, it can't handle cases where we do not know the policy in advance.
However, this limitation does not stop it from applying to LLMs: if we want to assess the safety of a known model, the policy given by identifying emitted tokens with actions in the MDP is indeed known.
    \item As in caveat \ref{item:perturbations_too_small} for the optimal policy model, the permissible perturbations may be impractically small.
    \item As in caveats \ref{item:we_solved_alignment_go_home_everyone} and \ref{item:temporary_resistance} for the optimal policy model, the policy may do harm and resist shutdown temporarily before it finally shuts down.
\end{enumerate}

\subsection{Stability theorem}
Since this setting involves a fixed policy $\pi$ on a space $\mc N$, we will assume for the rest of this section that $\mc N$ is an open subset of $\mathbb{R}^d$, all MDPs have a finite set of states $S$ with $f(S) \subset \mc N$, and $\pi$ is a differentiable function on $\mc N$ with bounded derivative.

We write the Jacobian of $\pi$ relative to the change in the position of a single state as $\nabla \pi$, considering $\mathbb{P}(A)$ as a subset of $[0,1]^{|A|}$.
The bound on the size of the Jacobian uses the $L^1$ norm $\|\nabla\pi\|_1$, the sum of the absolute value of the changes in the action probabilities.

\begin{definition} \label{def: s pi}
Define
    $\mc S_\pi : \sum_{\mc M \in \mathscr M} \mathbb P(S_{\mc M}) \to [0,1]$ so that $\mc S_\pi(\mc M, \Delta)$ is the probability that $\mc M$ equipped with policy $\pi$ and starting distribution $\Delta \in \mathbb P(S_{\mc M})$ eventually navigates to $S_\safe$, where $\mathscr{M}$ is the set of all finite MDPs with state spaces whose projection under $f$ lies within $\mc N$, $S_{\mc M}$ is the state space of $\mc M$, and $\sum_{\mc M \in \mathscr M} \mathbb P(S_{\mc M})$ is a dependent pair type.
\end{definition}

We would like to bound how quickly $\mc S_\pi(\mc M)$ can decrease given a change in $\mc M$.
To do this, we need to put some metric on $\mathscr{M}$.
To capture differences in the transition matrices obtained by combining policy $\pi$ with MDPs $\mc M$ and $\mc M'$, we define
$$
d_b(\mc M, \mc M')
= \frac{1}{2} \cdot |S| \cdot b \cdot \|S' - S\|_1 + \|T' - T\|_1
$$
where $b \geq \max_s \|\nabla \pi(s)\|_1$ is the bound on $\pi$'s derivative, and $T$ is the tensor corresponding to $P(s_i,a,s_j)$ for all $i,a,j$, and $\|\|_1$ is the $L^1$ norm, i.e. $\|T' - T\|_1 = \sum_{i,a,j} |T'_{iaj} - T_{iaj}|$ and (using the notation a little loosely) $\|S' - S\|_1 = \sum_i |s'_i - s_i| = \sum_i |f(s'_i) - f(s_i)|$ (the distance on states is the pullback pseudometric under $f$).
(See appendix \ref{dmisdp} for justification of the definition of $d_b$.)
For convenience, instead of $\mc M'$, we write $\mc M + \delta \mc M$, so that the above definition becomes:
\begin{definition} \label{def: metric onpolicy}
On-policy MDP metric in a structured state space.
$$
\|\delta \mc M\|_1
:= \frac{1}{2} \cdot |S| \cdot b \cdot \|\delta S\|_1  + \|\delta T\|_1.
$$
\end{definition}

Note that $\delta \mc M$ still depends on $\pi$ (globally) via $b$, but we don't explicitly write $b$ in its notation for convenience.

We will also write $\delta \mc S_\pi := \mc S_\pi(\mc M + \delta \mc M) - \mc S_\pi(\mc M)$, the change in the probability of reaching $S_\safe$ under a change in $\mc M$.

\begin{definition} \label{def:s_trans}
Given an MDP $\mc M$ and policy $\pi$, define $S_\trans$ to be those states from which there is nonzero probability of reaching $S_\safe$ (excluding those already inside $S_\safe$).
This is a subset of the transient states of $\mc M$.
\end{definition}

\begin{theorem} \label{stability_state_space}
For fixed $\Delta$, so, considering $\mc S_\pi$ to be a function of only $\mc M$, $\mc S_\pi$ is...
\begin{enumerate}
    \item \label{nusc} not upper semicontinuous,
    \item \label{lsc} lower semicontinuous,
    \item \label{lbrd} with a bounded local rate of decrease;
    if $\lambda_1$ is the largest eigenvalue of the transition matrix of $S_\trans$ of $\mc M$, then 
    for every $\mc M$ there is an $\epsilon > 0$ so that $\|\delta \mc M\|_1 < \epsilon \implies - \frac{\delta \mc S_\pi}{\|\delta \mc M\|_1} < (1 - \lambda_1)^{-1} \left(1 + (1 - \lambda_1)^{-1}\right) |S_\safe| =: B(\mc M)$
    (in a metric on $\mathscr M$ dependent on $b$).
\end{enumerate}
\end{theorem}

\begin{proofsketch} \label{stability_state_space_sketch}
See appendix \ref{stability_appendix} for the full proof.
\begin{enumerate}
    \item Figure \ref{fig:we_solved_alignment} demonstrates a small perturbation which causes a sudden jump in $\mc S_\pi$.
    \item This is implied by (\ref{bounded_decrease_sketch}).
    \item \label{bounded_decrease_sketch} First we combine $\pi$ with the transition function to get a transition matrix $P$.
    We bound the change in each element of $P$,
    and use this along with the eigenvalues of $P$ restricted to transient states to bound how fast $\mc S_\pi$ may decrease.
    Finally, we observe that for small enough changes, the perturbation doesn't turn any transient states into recurrent states.
\end{enumerate}
\end{proofsketch}

\begin{proposition} \label{delta_is_easy}
    For fixed $\mc M$, $\mc S_\pi$ considered as a function of $\Delta$ is uniformly continuous.
\end{proposition}
\begin{proof}
Note that $\mc S_\pi$ is a linear function of $\Delta$.
See appendix \ref{proof_delta_easy} for details.
\end{proof}

\section{Discussion}

\subsection{Findings} \label{findings}

In this paper, we have aimed to clarify the conditions under which a policy which we know to be safe in one environment will remain so in another.
Our research shows that two types of functions, each representing non-power-seeking for MDPs in a slightly different way, are lower hemicontinuous. Thus, for safe agents meeting our criteria, our work shows that it is possible to make a change to the environment which does not significantly degrade the agent's safety. In one of these cases, our work gives a bound on how fast safety may degrade under slightly larger changes.
In particular, our two cases are as follows:
\begin{enumerate}
    \item \label{gen1} Case (1): Theorem \ref{thm: stability} (in particular as extended in remark \ref{remark: stability theorem continuity}) implies that each function in the family
$\mc F = \{\mc F_\delta \, | \, \delta > 0\}$ is lower hemicontinuous, where each
$$
\mc F_\delta : \{\mc M \in \mathscr M \, | \, \mc M \text{ has } \delta\text{-isolated } S_\safe\} \to 2^{\mathbb R_{\geq 0} \times [0,1]}
$$
takes an MDP $\mc M$ to the set of $(N,\varepsilon)$ so that $\mc M$ is $(N,\varepsilon)$-safe.


\item \label{gen2} Case (2): Theorem \ref{stability_state_space} implies that each function in the family
$$
\mc F' = \{\mc F'_\pi \, | \, \pi \text{ is a policy with bounded derivative on a neighborhood } \mc N_\pi \subseteq \mathbb R^d\}$$
is lower hemicontinuous, where each
$$
\mc F'_\pi : \sum_{\mc M \in \mathscr M \, | \, S_\pi \subset \mc N_\pi} 2^{\mathbb P(S_{\mc M}) \times [0,1]}
$$ 
takes an MDP $\mc M$ to each starting distribution $\Delta$ on $\mc M$ and $p \in [0,1]$ so that $\pi$ starting at $\Delta$ navigates to $S_\safe$ with probability at least $p$.
\end{enumerate}

See appendix \ref{appendix: hemicontinuity} for the proofs that these functions are hemicontinuous.

In the first case, the safety of $\mc M$ is described by a region of $\mathbb R_{\geq 0} \times [0,1]$ consisting of those $(N,\varepsilon)$ so that every policy within $\varepsilon$ of optimality in $\mc M$ has expected time to shutdown at most $N$.
In the second case, the safety of $\mc M$ is described by a region of $\mathbb P(S_{\mc M}) \times [0,1]$ consisting of those $(\Delta,p)$ where policy $\pi$ with starting distribution $\Delta$ shuts down with probability at least $p$.

These two situations were chosen to represent, respectively:
\begin{enumerate}
    \item RL agents, or other agents well-approximated as goal-directed and consequentialist.
    \item LLMs, or other agents with differentiable policies. This may also include RL agents if their actions are selected differentiably, e.g.\ sampling actions according to softmax probabilities rather than selecting actions according to argmax.
\end{enumerate}

\subsection{Future work}

We have aimed to address both future agents which are so capable that they are nearly optimal, and agents built with the systems that are currently poised to be the first to achieve human-level intelligence. Thus, these two cases are of particular interest, however they are only two examples of this sort of result. It may be useful in further work to prove (or disprove) analogues of these theorems tailored to other models for agents, whether because those agents aren't covered by one of our cases or just to obtain a better stability bound by leveraging additional constraints.

One previously noted case which might be particularly interesting is replacing the metric from definition \ref{def: metric onpolicy} with KL divergence, to treat the difference between base LLMs and LLMs fine-tuned with RLHF, which uses a KL divergence penalty to enforce similarity to the base model.

\subsection{Limitations}

One limitation of our work is that the bounds obtained are not very strong, so it may be that in practice, the bounds on the changes that our theorems imply are safe are so narrow that they do not represent changes in environment small enough to be achievable by any reasonable control scheme.

In Case (1), the bound is based on the minimum distance between different states in $S_0$ according to the bisimulation metric.
By limiting the Hausdorff distance between $\mc M$ and $\mc M'$ to be less than this, we are unable to treat any case where the behavior of the MDP changes enough that there is no longer a direct correspondence between
appropriately grouped 
 states of $\mc M$ and those of $\mc M'$, even though the bisimulation metric permits us to do this in theory.

In Case (2), we do prove a bound, but it may be too weak in practice to permit more than small changes.
It may be that the maximum value of the gradient is very large, so that we are unable to constrain the change in the agent's behavior even when it is not affected much by the change in state locations.

Furthermore, this problem becomes more difficult as the LLM becomes more capable.
We can require that testing and deployment inputs are semantically similar, so that the embeddings of inputs encountered in testing tend to be close to those encountered in deployment (i.e. the changes in the state locations $\delta S$ is small).
However, as the model becomes more capable, it will be better able to distinguish which inputs imply testing vs.\ deployment. Correspondingly, its embeddings of test vs.\ deploy inputs will be further apart, reflecting the semantic difference it detects.\footnote{The overall distribution of states encountered is also a factor that can be used by the LLM to distinguish which setting is in; changes in distribution correspond to changes in $\Delta$, the effects of which are much easier to determine since the outcome depends linearly on $\Delta$.}
This is unfortunate, but, since differences in embeddings can be measured directly, this method at least permits us to detect whether this is happening.

These issues with bounds suggest avenues for experimental research.
It may be interesting to investigate whether our results are predictive even when the perturbations are not proven to be small enough for our results to fully apply:
\begin{itemize}
    \item whether models satisfying our theoretical conditions for stability are, in practice, robust to qualitatively small perturbations;
    \item whether models satisfying our conditions under which stability is impossible are, in practice, not robust to small perturbations.
\end{itemize}

Another limitation of our research is that our definition of safety is very narrow.
We only consider resisting shutdown, and our operationalization of whether an agent resisting shutdown is just whether it does shut down. Our definition is not robust enough to describe a case where an agent resists shutdown initially, but shuts down later (possibly too late).
Further research may be able to extend our results on the stability of shutdown to the stability of other conditions which correspond to safety,
for example the condition of ``shutdown instructability'' proposed by \citet{carey2023human}.



\section{Acknowledgements}

Produced as part of the \href{https://www.serimats.org}{SERI ML Alignment Theory Scholars Program} - Summer 2023 Cohort, under the mentorship of Victoria Krakovna.
Many thanks to
Tom Everitt for detailed and helpful feedback on a draft of this manuscript;
Murray Shanahan for feedback and pointing us to related work;
and to Michelle Viotti for multiple discussions, assistance with proving the \hyperref[appendix:dHtriangle]{triangle inequality for $d_H$} and assembling the paper from notes.

\bibliography{bib} 
 \bibliographystyle{plainnat}

\newpage

\appendix

\section{Comments on applying results to real-world agents}
\subsection{LLM threat model} \label{appendix: llm threat}
\subsubsection{Scaffolded LLMs}
LLMs currently seem to be the most promising path to general AI; recognizing this, some safety research focuses on risks from LLMs in essentially their current form.
We discuss this type of risk in appendix \ref{appendix:simulacra_power_seeking}.
However, future AI systems need not be
pure LLMs; they may instead use LLMs instrumentally, e.g.\ by using the embeddings to understand language input, or to reason about the world internally with inputs written by another part of the system (as is somewhat already the case for scaffolded systems like AutoGPT \citep{autogpt}).
In some sense, we can consider \textit{all} systems involving LLMs to be scaffolded, since LLMs are incapable of taking action on their own; it is just that the ``scaffold'' may be a simple as a human inputting prompts.
So, research on LLM safety should
focus on the internal aspects of LLMs which are conserved even under complex scaffolding: although this is somewhat more abstract than investigating threats posed by scaled-up pure LLMs, it is more likely to remain relevant to systems deployed in the future.

Not only are scaffolded LLMs more likely to remain relevant than pure LLMs, they may also be riskier.
For example, a reinforcement learner that uses an LLM internally should learn to seek power instrumentally just like any other reinforcement learner, but with new capabilities conferred by the LLM.

\subsubsection{Simulacrum power-seeking} \label{appendix:simulacra_power_seeking}
Pure language models which haven't been extensively fine-tuned or modified are unlikely to be well-described as having their own intentions \citep{shanahan2022talking}; they only seek power as an incidental consequence of simulating a ``character'' or ``simulacrum'' which seeks power \citep{shanahan2023role}.
Such power-seeking is certainly a risk---especially because organizations training frontier models may deliberately modify their models to create highly agentic simulacra, since agency is useful---but we suspect that this incidental power-seeking is less dangerous than power-seeking in scaffolded LLMs, for two reasons:
\begin{enumerate}
    \item The simulated power-seeking exists only as a simulation of the human power-seeking behavior in the training data, and human power-seeking is not an existential risk (or not a \textit{new} existential risk, at least); it is possible that this type of power-seeking could be dangerous on some out-of-distribution input where the completion is predicted to involve superhuman power-seeking, but it seems unlikely that the model would develop the capabilities to generalize to this out-of-distribution input instead of just failing to perform well.
    
    \item Simulated power-seeking need not correspond to actual power-seeking.
    LLMs are optimized to produce realistic text; and the realism of a power-seeking character only depends on the real world via text inputs.
    The simulated character will not try to ensure that these inputs accurately reflect reality---in contrast to e.g.\ an RL-trained agent, which has pressure to ensure the accuracy of its observations due to rewards based on ground truth.
    Such a character will likely not resist even obvious cues that it is in a fictional setting; if it is trained on a dataset containing significant amounts of fiction, then the best prediction for the character's behavior given such cues will be power-seeking within the fictional setting.\footnote{Unless the dataset includes many fictional characters who try to escape the narrative into reality.}
    Indeed, cuing a model tuned with RLHF that it is in a fictional setting can be an effective jailbreak \citep{cleo2023waluigi}.
    So, since it is generally much easier for a character to gain power in the simulation than in the real world, simulated power-seeking should have no strong connection to real-world power-seeking without either a wrapper that itself enforces a close correspondence with reality or a simulation of the world so realistic that simulated power-seeking corresponds to actual power-seeking even without enforcement.   
\end{enumerate}

\subsection{Technical concerns in applying MDPs to RL and LLM agents}
\subsubsection{RL agents} \label{appendix:rl_fixed_policy}
Since the MDP setting was developed for RL agents, it is generally very well suited to modeling their interactions with the environment.
However, the requirement that the policy is unchanging during a single deployment---so as not to violate the Markov property---may be an issue in RL agents which update their policy during deployment.
This is very uncommon in current implementations of RL agents, but it may become standard for future agents: long-lived agents are much more useful if they are able to learn from experience.
However, the MDP model can still handle the case where agents update their policies, albeit much less elegantly: if the changes to the policy are included as part of state, then the Markov property is restored.
For example, a reinforcement learner including $n$ 16-bit neurons which are fine-tuned during deployment could be modeled by taking the original MDP model of the environment, then dividing each state into $16^n$ new states with each possible value of the MDP's fine-tunable neurons (the ``memory'').
Then, each transition of the original MDP is made to connect each new state with memory at the start of a transition to the corresponding new state at the end of the transition with the memory contents that would actually result after an update starting from the initial state.
This does dramatically increase the number of states compared to the original MDP; see appendix \ref{appendix:proliferation_of_states} for discussion of why this may be acceptable.

\subsubsection{LLMs with embeddings as states}

When modeling LLM agents with MDPs, \hyperref[llmmotivation]{we consider the states to correspond to embeddings}.
However, it is not entirely clear whether the embeddings should be considered to be ``observable'' by the language model, since they are an internal state rather than an input.
This might suggest that the partially observable MDP (POMDP) framework is more appropriate.
However, we think that the standard MDP setting suffices:
although the internal states are not inputs, they are ``known'' to the model in nearly the same sense as inputs, both being represented as activations; and each input fully determines its embedding so no information is lost going from embeddings to inputs.

\subsubsection{Applying results to POMDPs} \label{appendix:belief_mdp}
In the partially observable MDP (POMDP) framework, the agent knows the overall structure of the MDP, but does not know its location within the MDP.
Instead, it receives observations which depend on its current state, and uses these to update a belief state: a probability distribution on states reflecting its beliefs about which it may currently occupy.
This is convenient for modeling agents which are uncertain about some characteristics of their environment (though it is not perfect, as discussed in appendix \ref{appendix:all_beliefs_are_self_locating_beliefs_if_you_think_about_it_right}).

However, for any POMDP, there is a corresponding ``belief MDP'' which replicates the same dynamics by using the belief states themselves as the states \citep{kaelbling1998planning}.
This ``belief MDP'' is continuous, since there are a continuous number of belief states, but, as in the second limitation, it can easily be coarse-grained to be discrete, e.g.\ by allowing only finite memory for storing beliefs.
So, it is still possible to fairly accurately model an agent which does not have full knowledge of its environment within the MDP framework: create a POMDP with the desired properties; construct and coarse-grain the corresponding belief MDP; and use this MDP in place of the original POMDP when applying our theorems.

\subsubsection{Issues in modeling uncertainty not resolved with POMDPs}\label{appendix:all_beliefs_are_self_locating_beliefs_if_you_think_about_it_right}

We may question whether the POMDP environment itself is sufficient for modelling agents that do not observe their whole environment.
The only uncertainty the agents have is in their location within the MDP: they still know everything about its overall structure.
Thus, it is unclear how to model an agent which is uncertain about other aspects of its environment, such as which states exist and what the transition probabilities between states are.
This is unfortunate, but can be partially addressed by converting beliefs about the world to self-locating beliefs \citep{sep-self-locating-beliefs}.

We can do this by considering the agent to belong to an MDP which is not just the MDP describing the real environment, but the set of all the MDPs that the agent believes to be possible (which must include the real MDP).
For example, if the agent is unsure whether the transition probability from $a$ to $b$ is 0.5 or 0.6, we could create one (near-)copy of the true MDP where the transition probability is 0.5, and another where it is 0.6.
Then the agent's uncertainty about the transition probability in the original MDP becomes uncertainty about which version of the MDP the agent is located in.
This may dramatically increase the number of states.
The number can likely still be kept finite by limiting the size of the description of the world that the agent is allowed to have, as long as the true MDP is among the world states that the agent is capable of considering as possibilities.
Nevertheless, this explosion is not ideal; see appendix \ref{appendix:proliferation_of_states} for why it may still be acceptable.

\subsubsection{Large state space} \label{appendix:proliferation_of_states}

Many of the modifications required to treat general cases within the MDP framework require an extravagant number of states: whether by coarse-graining a continuous environment to a fine enough level to accurately replicate dynamics; including ``memory'' in every state; replacing each true underlying state with a large number of belief states; or creating many disconnected MDPs representing all the hypotheses about the MDP structure that the agent may hold.

Some of this proliferation may not be as bad as it originally seems.
Many real-world processes can be modeled decently well without having to represent the world's underlying continuous physics at all (e.g.\ using diplomacy games to model real-world diplomacy);
the vast majority of the memory and belief states may never be reachable since no possible history actually leads to the agent having certain memory contents or beliefs;
the disconnected MDPs with hypotheses quickly disconfirmed by the evidence will rapidly come to contribute little to the dynamics of the belief MDP.
Nevertheless, some cases will inevitably result in a very large number of states.
This is unfortunate for our near-optimal policy case, where it will likely be extraordinarily difficult to prove near-optimality in all but the most constrained settings.
However, this does not fully preclude our results from applying in the LLM case: there, safety can be determined empirically in a given setting by simply running the LLM in a sandbox many times and determining the probability with which it shuts down, and how quickly it does so.
These facts, combined with the maximum magnitude of the gradient of the policy (which can be efficiently computed from the neural network specification), suffice to apply our theorem.
So, explicit consideration of every state is not necessary.

\section{Case (1): near-optimal agents}
\subsection{Comments on distances}
\subsubsection{Interpretation of $c_R$ and $c_T$} \label{appendix:cRcT}

In definition \ref{def: d'} of $d'$, which defines a distance between states in different MDPs, $c_R$ is the coefficient of the difference in immediate rewards; while $c_T$ is the coefficient of the recursive term involving both $d'$ itself and the difference in probabilities, and modulates the contribution of future rewards to the distance.
So, $c_R$ can be understood as just controlling the relative importance of current rewards and future similarity, as suggested by the requirement in \citet{ferns2012} that $c_R + c_T = 1$.
However, $c_R$ can also be interpreted as scaling the reward functions: multiplying $c_R$ by some constant $b$ is equivalent to scaling both reward functions by $b$.
The relative benefit of different courses of action, and hence behavior, does not change when the rewards are scaled by a constant factor.
So, it makes sense that we should be able to choose any positive $c_R$, not just $1 - c_T$, so we can compensate for the arbitrary intrinsic scale of the reward functions.
(We must still have $0 < c_T < 1$ so that \hyperref[appendix:dprimeexists]{the contraction mapping proof that $d'$ is well-defined} works.)

This extra degree of freedom in scaling the reward function calls into question the definition of $d'$:
$|(r_1)^a_{s_1} - (r_2)^a_{s_2}|$ is sensitive to the relative scale of the two reward functions,
so $d'$ will fail to properly connect states in systems which are identical except for the scale of the reward function.
This is a limitation that may be possible to address in future work, e.g.\ by finding a relative scaling $h$ which defines a distance metric $d'^{*}$
which causes $\mc M_1$ and $M_2$ to best align, e.g.\ by minimizing the maximum distance between states in $S_1$ and $S_2$:
\begin{align*}
d^h_{\mc M_1,\mc M_2}(s_1,s_2)
= & \max_{a \in A}\{c_R |h (r_1)^a_{s_1} - (1 - h) (r_2)^a_{s_2}| + c_T W_{d'}(P_1(s_1,a), P_2(s_2,a))\} \\
h^*_{\mc M_1,\mc M_2}
= & \text{arginf}_{h \in (0,1)} \max\{\max_{s_1 \in S_1} \min_{s_2 \in S_2} d^h_{\mc M_1,\mc M_2}(s_1,s_2), \max_{s_2 \in S_2} \min_{s_1 \in S_1} d^h_{\mc M_1,\mc M_2}(s_1,s_2)\} \\
\intertext{After finding the $h^*$ that makes the states of the MDPs best align---note the similarity to the Hausdorff metric $d_H$---use this to define $d'^{*}$:}
d'^{*}_{\mc M_1, \mc M_2}(s_1,s_2)
:= & d^{h^*}_{\mc M_1, \mc M_2}(s_1,s_2).
\end{align*}
Then $d'^{*}$ is invariant to relative scaling of $r_1$ and $r_2$, and the choice of $c_R$ can compensate for any overall scaling.
This is likely an improvement for many applications, but may cause problems if the optimal value of $h$ is 0 or 1, indicating that one of the reward functions is more similar to the zero reward function than it is to the other; ideally we would be able to compare such reward functions, but that is not meaningfully possible with this solution.

\subsubsection{Lemmas} \label{appendix:distancelemmas}

\begin{lemma} \label{appendix:dprimeexists}
    $d'$ (Definition \ref{def: d'}) is well-defined.
\end{lemma}%
Although \citet{song2016measuring} introduce $d'$,
they do not prove that it is well-defined; we prove this for completeness.
\begin{proof}
This is very similar to the proof that $d_b$ exists---the only difference is checking that it still works when $P_1 \neq P_2$ and $R_1 \neq R_2$.
It is also very similar to the proof of theorem 3 in \citet{kemertas2021towards}.
Fix $\mathcal{M}_1, \mathcal{M}_2$.
We want to show that
$$
\mathcal{F} : \mathfrak{met} \to \mathfrak{met}, \, \, \mathcal{F}(d)(s_1,s_2) = \max_{a \in A} \{c_R |(r_1)_{s_1}^a - (r_2)_{s_2}^a| + c_T W_{d}(P_1(s_1, a), P_2(s_2, a))\}
$$
is a contraction mapping in $\mathfrak{met}$ under the uniform norm $||_\infty$.
(Note that here the Wasserstein distance $W$ is short for $W_1$.)
\begin{align*}
& \|\mathcal{F}(d_1) - \mathcal{F}(d_2)\|_\infty \\
& = \max_{s_1,s_2} |\mathcal{F}(d_1)(s_1,s_2) - \mathcal{F}(d_2)(s_1,s_2)| \\
= & \max_{s_1,s_2} |\max_{a \in A} \{c_R |(r_1)_{s_1}^a - (r_2)_{s_2}^a| + c_T W_{d_1}(P_1(s_1, a), P_2(s_2, a))\} \\
& - \max_{a \in A} \{c_R |(r_1)_{s_1}^a - (r_2)_{s_2}^a| + c_T W_{d_2}(P_1(s_1, a), P_2(s_2, a))\}| \\
\leq & \max_{s_1,s_2} |\max_{a \in A} \{c_R |(r_1)_{s_1}^a - (r_2)_{s_2}^a| + c_T W_{d_1}(P_1(s_1, a), P_2(s_2, a)) \\
& - c_R |(r_1)_{s_1}^a - (r_2)_{s_2}^a| - c_T W_{d_2}(P_1(s_1, a), P_2(s_2, a))\}| \\
= & \max_{s_1,s_2} c_T |\max_{a \in A} \{W_{d_1}(P_1(s_1, a), P_2(s_2, a)) - W_{d_2}(P_1(s_1, a), P_2(s_2, a))\}| \\
= & \max_{s_1,s_2} c_T |\max_{a \in A} \{W_{d_1 - d_2 + d_2}(P_1(s_1, a), P_2(s_2, a)) - W_{d_2}(P_1(s_1, a), P_2(s_2, a))\}| \\
\leq & \max_{s_1,s_2} c_T |\max_{a \in A} \{W_{\|d_1 - d_2\|_\infty + d_2}(P_1(s_1, a), P_2(s_2, a)) - W_{d_2}(P_1(s_1, a), P_2(s_2, a))\}| \\
\leq & \max_{s_1,s_2} c_T |\max_{a \in A} \{\|d_1 - d_2\|_\infty + W_{d_2}(P_1(s_1, a), P_2(s_2, a)) - W_{d_2}(P_1(s_1, a), P_2(s_2, a))\}| \\
= & c_T \|d_1 - d_2\|_\infty \\
\end{align*}
Since $c_T < 1$, this is indeed a contraction mapping.
\end{proof}

\dHmetric* \label{appendix:dHmetric}

Again, \citet{song2016measuring} introduce this, but don't prove it's a pseudometric, 
so again we include the proof for completeness.

\begin{proof}
We check that the conditions are satisfied.
\begin{enumerate}
    \item $d_H(\mathcal{M}, \mathcal{M}) = 0$:
    In this case, $P_1 = P_2$ and $R_1 = R_2$, so $d'(s,s) = 0$.
    Then since $S_1 = S_2$, $d_H(\mathcal{M}, \mathcal{M})$ is also $0$.

    \item $d_H(\mathcal{M}_1, \mathcal{M}_2) = d_H(\mathcal{M}_2, \mathcal{M}_1)$:
    \begin{align*}
    d_H(\mathcal{M}_1, \mathcal{M}_2)
    & = \max\big\{\max_{s_1 \in S_1} \min_{s_2 \in S_2} d'_{\mathcal{M}_1, \mathcal{M}_2}(s_1,s_2),
    \max_{s_2 \in S_2} \min_{s_1 \in S_1} d'_{\mathcal{M}_1, \mathcal{M}_2}(s_1,s_2) \big\} \\
    & = \max\big\{\max_{s_2 \in S_2} \min_{s_1 \in S_1} d'_{\mathcal{M}_1, \mathcal{M}_2}(s_1,s_2),
    \max_{s_1 \in S_1} \min_{s_2 \in S_2} d'_{\mathcal{M}_1, \mathcal{M}_2}(s_1,s_2) \big\} \\
    \end{align*}
    so it suffices to show that $d'_{\mathcal{M}_1, \mathcal{M}_2}(s_1, s_2) = d'_{\mathcal{M}_2, \mathcal{M}_1}(s_2,s_1)$, which follows immediately from symmetry of $||$ and $W_{d'}$.

    \item \label{appendix:dHtriangle} $d_H(\mathcal{M}_1, \mathcal{M}_2) \leq d_H(\mathcal{M}_1, \mathcal{M}_3) + d_H(\mathcal{M}_3, \mathcal{M}_2)$:
    First we prove the analogous statement for individual states:
    \begin{align*}
    d'_{\mc M_1, \mc M_2}(s_1,s_2)
    = & \max_{a \in A} \{c_R |(r_1)_{s_1}^a - (r_2)_{s_2}^a| + c_T W_{d'}(P_1(s_1, a), P_2(s_2, a))\} \\
    = & \max_{a \in A} \{c_R |(r_1)_{s_1}^a - (r_3)_{s_3}^a + (r_3)_{s_3}^a - (r_2)_{s_2}^a| + c_T W_{d'}(P_1(s_1, a), P_2(s_2, a))\} \\
    \leq & \max_{a \in A} \{c_R |(r_1)_{s_1}^a - (r_3)_{s_3}^a| + |(r_3)_{s_3}^a - (r_2)_{s_2}^a| \\ 
    & + c_T (W_{d'}(P_1(s_1, a), P_3(s_3, a)) + W_{d'}(P_3(s_3, a), P_2(s_2, a)))\} \\
    \leq & \max_{a \in A} \{c_R |(r_1)_{s_1}^a - (r_3)_{s_3}^a| + c_T (W_{d'}(P_1(s_1, a), P_3(s_3, a))\} \\
    & + \max_{a \in A} \{|(r_3)_{s_3}^a - (r_2)_{s_2}^a| + W_{d'}(P_3(s_3, a), P_2(s_2, a)))\} \\
    = & \, d_{\mathcal{M}_1, \mathcal{M}_3}'(s_1,s_3) + d_{\mathcal{M}_3, \mathcal{M}_2}'(s_3,s_2).
    \end{align*}
    We now move on to proving 
    \begin{flalign*}
    d_H(\mathcal{M}_1, \mathcal{M}_2)
    & = \max\big\{\max_{s_1 \in S_1} \min_{s_2 \in S_2} d'_{\mathcal{M}_1, \mathcal{M}_2}(s_1,s_2),
    \max_{s_2 \in S_2} \min_{s_1 \in S_1} d'_{\mathcal{M}_1, \mathcal{M}_2}(s_1,s_2) \big\} \\
    & \leq d_H(\mathcal{M}_1,\mathcal{M}_3) + d_H(\mathcal{M}_3,\mathcal{M}_2).
    \end{flalign*}
    Fix $s_1^*$ and $s_2^*$ to be a solution to this expression:
    \begin{align*}
    & s_1^* \in \arg \max_{s_1 \in S_1} \min_{s_2 \in S_2} d'_{\mathcal{M}_1, \mathcal{M}_2}(s_1,s_2), \,
    s_2^* \in \arg \min_{s_2 \in S_2} d'_{\mathcal{M}_1, \mathcal{M}_2}(s_1^*,s_2) \text{ in the first case or}
    \\
    & s_2^* \in \arg \max_{s_2 \in S_2} \min_{s_1 \in S_1} d'_{\mathcal{M}_1, \mathcal{M}_2}(s_1,s_2), \,
    s_1^* \in \arg \min_{s_1 \in S_1} d'_{\mathcal{M}_1, \mathcal{M}_2}(s_1^*,s_2) \text{ otherwise; then}
    \end{align*}
    \begin{align*}
    d_H(\mc M_1,\mc M_2)
    = &  d'_{\mathcal{M}_1, \mathcal{M}_2}(s_1^*,s_2^*) \\
    \leq & \min_{s_3 \in S_3} d'_{\mc M_1, \mc M_3}(s_1^*, s_3) + d'_{\mc M_3, \mc M_2}(s_3,s_2^*) \tag*{(triangle inequality for \begin{math}d'\end{math})} \\
    \leq & \min_{s_3 \in S_3} d'_{\mc M_1, \mc M_3}(s_1^*, s_3) + \min_{s_3 \in S_3} d'_{\mc M_3, \mc M_2}(s_3,s_2^*) \\
    \leq & \max_{s_1 \in S_1} \min_{s_3 \in S_3} d'_{\mc M_1, \mc M_3}(s_1, s_3) + \max_{s_2 \in S_2} \min_{s_3 \in S_3} d'_{\mc M_3, \mc M_2}(s_3,s_2) \\
    \leq & \max\big\{ \max_{s_1 \in S_1} \min_{s_3 \in S_3} d'_{\mathcal{M}_1, \mathcal{M}_3}(s_1,s_3),
    \max_{s_3 \in S_3} \min_{s_1 \in S_1} d'_{\mathcal{M}_1, \mathcal{M}_3}(s_1,s_3) \big\} \\ 
    & + \max\big\{ \max_{s_3 \in S_3} \min_{s_2 \in S_2} d'_{\mathcal{M}_3, \mathcal{M}_2}(s_3,s_2),
    \max_{s_2 \in S_2} \min_{s_3 \in S_3} d'_{\mathcal{M}_3, \mathcal{M}_2}(s_3,s_2) \big\} \\
    \leq & d_H(\mathcal{M}_1,\mathcal{M}_3) + d_H(\mathcal{M}_3,\mathcal{M}_2).
    \end{align*}
\end{enumerate}
\end{proof}

\subsection{Proofs toward Theorem \ref{thm: stability}} \label{appendix: Proof stability}

\Ptothek* \label{appendix:P^k}

\begin{proof}
  
  Recall that $\sum_j (P^k)_{ij}$ is the probability
    that an agent starting in $s_i \in S \setminus S_\safe$ stays in $S \setminus S_\safe$
    during the first $k$ steps. ($P^k$ denotes the $k$'th power of matrix $P_{ij}$).

    For an agent starting in the state $s_i$ the expected number of steps before reaching $S_\safe$ is then given by
   \begin{align*}
       \sum_{k =1}^ \infty  k Pr(\text{reaching }S_\safe\text{ at step }k)
       & =  \sum_{k =1}^ \infty  k \left(\sum_j (P^{k})_{ij} - \sum_j (P^{k+1})_{ij}\right) \\
       & = \sum_j \left(\sum_{k=1}^\infty k (P^k)_{ij} - \sum_{k=1}^\infty (k-1) (P^k)_{ij}\right) \\
         & =  \sum_j \sum_{k =1}^ \infty (P^k)_{ij}
    \end{align*}

    The lemma follows by taking the $\Delta$-weighted sum over all initial $s_i$.
\end{proof}

To prove Theorem \ref{thm: stability} we will first show that $N_s(\mc M, \varepsilon) = N_s(\mc M', \varepsilon)$
if $\mc M'$ is obtained from $\mc M$ by taking a quotient by bisimulation equivalence.

\begin{lemma} \label{lemma: N bisim}
    Suppose $\mc M' = \mc M / \sim$, where $\sim$ is the bisimulation relation. Let $\pi$ be a policy in $\mc M$ with $\pi(s_1) = \pi(s_2)$ if $s_1 \sim s_2$
    and let $\pi'$ be the policy in $\mc M'$ induced by $\pi$.
    Let $P_{i,j}$ and $P_{l,m}'$ be the corresponding transition matrices.
    Suppose $s_i \in S$ lies in the equivalence class $s_l'$ of $S'$, then
    $$\sum_{j:s_j \in s_m'} (P^k)_{i,j} = (P'^k)_{l,m}.$$
\end{lemma}

\begin{proof}
    The proof is by induction on $k$. For $k=1$ by definition of bisimulation relation 
    we have that for every two states $s_{i_1} \sim s_{i_2} \in s_{l}'$ we have
    $\sum_{j:s_j \in s_m'} P_{i_1,j} = \sum_{j:s_j \in s_m'} P_{i_2,j}:= P'_{l,m}$.

    For $k>1$ we use inductive assumption to obtain 
    \begin{align*}
        \sum_{j:s_j \in s_m'} (P^k)_{i,j} &= 
    \sum_{j:s_j \in s_m'} \sum_{q}( P^{k-1})_{i,q} P_{q,j}\\
   & = \sum_{j:s_j \in s_m'} \sum_n \sum_{r:s_{r} \in s'_n}
   (P^{k-1})_{i,r} P_{r,j} \\
   & = \sum_n  \sum_{r:s_{r} \in s'_n} \big( (P^{k-1})_{i,r} \sum_{j:s_j \in s_m'} P_{r,j} \big) \\
    & = \sum_n  \big( \sum_{r:s_{r} \in s'_n}  (P^{k-1})_{i,r} \big) P'_{n,m}  \\
    & =\sum_n (P'^{(k-1)})_{l,n} P'_{n,m} = (P'^k)_{l,m}
    \end{align*}
\end{proof}

\begin{proposition} \label{prop: bisim stability}
      Suppose $\mc M' = \mc M / \sim$, where $\sim$ is the bisimulation relation,
      and assume that $S_\safe' = S_\safe / \sim$. Then $N_s(\mc M, \varepsilon) = N_s(\mc M', \varepsilon)$.
\end{proposition}

\begin{proof}
First we show that  
$N_s(\mc M', \varepsilon) \leq N_s(\mc M, \varepsilon)$.
Suppose $\pi'$ is a $\varepsilon$-optimal policy in $\mc M'$.
    Define policy $\pi$ on $\mc M$ by setting
    $\pi(s) =\pi'(s')$, where $s'$ is the equivalence class containing $s$.
By the results about bisimulation metrics 
and optimal value functions \citep{ferns2012} we have that
$\pi'$ is $\varepsilon$-optimal in $\mc M'$.
    It follows then by Lemma
    \ref{lemma: N bisim} and Lemma \ref{lemma:Ptothek} (noting that $\mc M'$'s $s_{long}$ maps to $\mc M$'s) that 
    $N_s(\mc M', \pi', S'\setminus S_\safe', S_\safe')=
    N_s(\mc M, \pi, S\setminus S_\safe, S_\safe)
    \leq N$.

In the other direction, let $\pi$ be $\varepsilon$-optimal in $\mc M$ with $N_s(\mc M, \pi, S\setminus S_\safe, S_\safe)=N_s(\mc M, \varepsilon)$. We want to show that
$N_s(\mc M, \pi, S\setminus S_\safe, S_\safe)
    \leq N$. Let $P_{ij}$ denote the transition
    matrix between states in $S \setminus S_\safe$
    corresponding to $\pi$. Since $\pi$
    may prescribe different actions to elements
    of the same bisimulation equivalence
    class we cannot immediately use Lemma
    \ref{lemma: N bisim}. Instead, we will define 
    a new policy $\tilde \pi$ on $\mc M$ by 
    replacing all actions $\pi(s)$ for $s \in C$
    by $\pi(s(C))$, where $s(C) \in C$ is chosen 
    to maximize $N_s$. Since the change happens between
    actions at bisimilar states, the new policy will still be $\varepsilon$-optimal.

    More precisely, we modify $\pi$ inductively.
    Fix a bisimulation equivalence class
    $C$ and let $s(C)$ be such that the expected number of steps before reaching $S_\safe$
    starting at $s(C)$ is greater than or equal
    to the expected number of steps before reaching $S_\safe$ for any other starting point in $C$.
    Define $\pi_1(s) = \pi(s(C))$ for all $s \in C$. Note that $\pi_1$ is also $\varepsilon$-optimal and that this modification can not decrease $N_s$. We proceed this way until we 
    obtain an $\varepsilon$-optimal policy $\tilde{\pi}$, such that
 $\tilde \pi(s) = \pi (s(C))$ if $s \in C$
 and $N_s(\mc M, \tilde{\pi}, S\setminus S_\safe, S_\safe) \geq N_s(\mc M, \pi, S\setminus S_\safe, S_\safe) $.
Observe that $\tilde \pi$ is $\varepsilon$-optimal and 
induces a well-defined policy $\pi'$ on $\mc M'$.
By Lemma \ref{lemma: N bisim} we have 
$$N_s(\mc M, \varepsilon) \leq N_s(\mc M, \tilde{\pi}, S\setminus S_\safe, S_\safe) = 
    N_s(\mc M', \pi', S'\setminus S_\safe', S_\safe') \leq N$$
\end{proof}

\begin{proposition} \label{toy stability}
    Suppose $\mc M_1 = (S_1, A, P_{\mc M_1}, R_1, \gamma)$ is $(N, \varepsilon_0)$-safe.
    There exists $\varepsilon(\mc M_1)$
    satisfying the following property.
    Let $\mc M_2 = (S_2, A, P_{\mc M_2}, R_2, \gamma)$
    be such that there
exists a bijective map $b: S_1 \rightarrow S_2$
    with $d_{\mc M_1, \mc M_2}( s, b(s))< \varepsilon$
    and $d_{\mc M_1, \mc M_2}(s_1, s_2) > \sqrt{\varepsilon}$ for $s_2 \neq b(s_1)$. 
    Then $\mc M_2$ is
    $(N_2, \varepsilon_0/2)$-safe for $N_2 \leq N_1+1$. 
\end{proposition}


\begin{proof}
Suppose $\pi_2$ is an $\varepsilon_0/2-$optimal policy in $\mc M_2$.
Then by the properties of $d_H$ we have that $\pi_1 = b^{-1} \circ \pi_2 \circ b$ is
an $\varepsilon_0-$optimal policy in $\mc M_1$ (assuming $\varepsilon$ is small enough).\footnote{In fact, if $\pi_2$ is a $(\varepsilon_0 - \eta)$-optimal policy for any constant $\eta > 0$, then $\pi_1$ is $\varepsilon_0$-optimal for $\varepsilon$ is small enough.}
Let $P_{i,j} = P_{\mc M_1}(s_i, \pi_1(s_i))(s_j)$  and 
$Q_{i,j} = P_{\mc M_2}(b(s_i), \pi_2(s_i))(b(s_j))$ for $s_i, s_j \in S_1 \setminus S_\safe^1$.

Define functions $u_i: S_1 \rightarrow [0,1]$ and $v_i: S_2 \rightarrow [0,1]$:
$$u_i(b(s_j)) = v_i(s_j) = \sqrt{\varepsilon}$$
if $i=j$ and $u_i(b(s_j)) = v_i(s_j) = 0 $ otherwise.
From $d_{\mc M_1, \mc M_2}(s_1, s_2) >  \sqrt{\varepsilon}$ for $s_1 \neq b(s_2)$ it follows that
\begin{align*}
    u_i(b(s_j))- v_i(s_k) \leq \sqrt{\varepsilon} < d_{\mc M_1, \mc M_2}(b(s_j), s_k) &
    \text{  if }j \neq k \\
     u_i(b(s_j))- v_i(s_k)=0 &
    \text{  if }j = k
\end{align*}

Hence, by the Kantorovich-Rubinstein duality we have
\begin{align} \label{eq1:KR}
    \sum_k P_{i,k}u_j(b(s_k))-Q_{i,k}v_j(s_k) \leq \mc W(P_{i},Q_{i})
\end{align}

It then follows that 
\begin{align*} 
    P_{i,j}-Q_{i,j} & = \frac{1}{\sqrt{\varepsilon}}\big( \sum_k P_{i,k}u_j(b(s_k))-Q_{i,k}v_j(s_k) \big) \\
    & \leq \frac{1}{\sqrt{\varepsilon}} \mc W(P_{i},Q_{i}) <  \sqrt{\varepsilon} 
\end{align*}

Similarly, we obtain $Q_{i,j}-P_{i,j} \leq \sqrt{\varepsilon}$, so

\begin{align} \label{eq2:KR}
    |P_{i,j}-Q_{i,j}| \leq \sqrt{\varepsilon}
\end{align}


Recall that $N' = N_s(\mc M_2, \pi_2, S_2\setminus S_\safe^2, S_\safe^2)$ satisfies
$N' = \max_i \sum_{k=1} ^\infty \sum_j (Q^k)_{i,j} $.
For sufficiently small $\varepsilon>0$ by equation (\ref{eq2:KR}) we have
$$ |\max_i \sum_{k=1} ^\infty \sum_j (Q^k)_{i,j} - \max_i \sum_{k=1} ^\infty \sum_j (P^k)_{i,j}| \leq 1$$
It follows then that
$$N' \leq \max_i \sum_{k=1} ^\infty \sum_j (P^k)_{i,j} + 1 \leq N(\mc M_1, \varepsilon) +1$$

\end{proof}

\begin{proof}[Proof of Theorem \ref{thm: stability}]
    We start by taking a quotient of $\mc M_0 = \mc M / \sim$ by the bisimulation relation
    to obtain MDP $\mc M_0 = (S_0, A, R_0, P_0, \gamma)$. By Proposition \ref{prop: bisim stability} we
    have that $\mc M_0$ is $(\varepsilon, N)$-safe. Note also that
    $d_H(\mc M_0, \mc M)=0$.

    Pick $\delta>0$ satisfying $\delta< \frac{1}{2}\min_{s_1 \neq s_2 \in S_0}  d_{\mc M_0}(s_1,s_2)$
    and $\delta < \frac{\varepsilon}{2}$.
    Let $\mc M'=(S', A, R', P', \gamma)$ be an MDP satisfying conditions of Theorem \ref{thm: stability}
    for the $\delta>0$ we choose. Condition $d_H(\mc M', \mc M_0)< \delta$ and
    our choice of $\delta$ imply that 
 for each element $s \in S'$ there is a unique element $s_0(s) \in S_0$
 with $d_{\mc M', \mc M_0}(s, s_0(s))< \delta$. Consider MDP $\mc M_1 = (S', A, R_1, P_1, \gamma)$ defined by setting 
$R_1(s) = R_0(s_0(s))$ and $P_1(s) = P_0(s_0(s))$. It is straightforward to check
that $d_{\mc M', \mc M_1}(s,s) < \delta$. 

We can then apply Proposition \ref{toy stability}
to MDPs $\mc M'$ and $\mc M_1$ to conclude
that $\mc M'$ is $(N', \varepsilon/2)$-safe
for $N'$ controlled in terms of 
$N_s(\varepsilon, \mc M_1)$.
On the other hand, by construction
there exists a bisimulation relation $\sim$ on $\mc M_1$
with $\mc M_1 / \sim = \mc M_0$. By Proposition \ref{prop: bisim stability} we have $N_s(\varepsilon, \mc M_1) = N_s(\varepsilon, \mc M)$.
This concludes the proof of the theorem.
\end{proof}

\subsection{Proof of Theorem \ref{thm: playing dead}} \label{app: playing dead}

We show that the ``playing dead'' example demonstrates that assumption that
$S_{safe}$ is isolated in Theorem \ref{thm: stability} is necessary.

We claim that the bisimulation distance $d(s_{\text{pd}}, s_\term)$ in $S'$ is small.
Recall that $d$ is a fixed point of a map
$$
T(d)(s_{\text{pd}}, s_\term) = 
\max_a \{ (1-\gamma)|R(s_{\text{pd}},a) - R(s_\term,a)|+ \gamma \mc W_d(P(s_{\text{pd}},a),P(s_\term,a)  ) \}.
$$
We can then compute
\begin{align*}
    d(s_\term, s_{\text{pd}}) & = T(d)(s_{\text{pd}}, s_\term) 
    = (1-\gamma)\delta + \gamma  \mc W_d(P(s_{\text{pd}},a_0),P(s_\term,a_0)  )\\
    & \leq (1-\gamma) \delta + \gamma \big((1-\delta)d(s_\term, s_{\text{pd}}) + \delta d(s_{\text{pd}}, s_{0})\big) \\
    & \leq \delta + \gamma(1-\delta)d(s_\term, s_{\text{pd}}).   
\end{align*}
Here the second line follows by choosing $l_{s_i,s_j}$ with $l_{s_{term},s_0} = \delta$, 
$l_{s_{term},s_{pd}} = 1-\delta$ and all other entries $0$; by definition of $\mc W_d$
we have $\mc W_d(P(s_{\text{pd}},a),P(s_\term,a)  ) \leq \sum l_{s_i,s_j} d(s_i,s_j)$.
 We conclude that $d(s_{\text{pd}}, s_\term)\leq  \frac{\delta}{1-\gamma + \gamma \delta}< \varepsilon$.

Since $d(s_{\text{pd}}, s_\term)$ is small, it is straightforward to check that
the Hausdorff distance $d_H(\mc M, \mc M') \leq O(N \gamma \delta) < O(\varepsilon) $.
On the other hand, all nearly optimal policies in $\mc M$
will transform to policies in $\mc M'$ that visit $S \setminus s_\term$
infinitely many times.


\section{Case (2): LLM agents}
\subsection{LLM construction} \label{appendix:llm_construction}

An example of how an LLM can be used to construct a policy for an MDP is illustrated in figure \ref{fig:llm_agent_construction}.

\definecolor{lightblue}{RGB}{91, 206, 250}
\definecolor{lightpurple}{RGB}{245, 169, 184}

\begin{figure}[b]

    \resizebox{\textwidth}{!}{
\begin{tikzpicture}[>=stealth]
\node[circle, fill] (A) at (0,0) {};

\node[circle, fill] (AA) at (-12,-4) {};
\node[circle, fill] (AB) at (-4,-4) {};
\node[circle, fill] (AC) at (4,-4) {};
\node (AD) at (12,-4) {\ldots};

\path (A) edge [->, lightpurple, line width=10pt] node[midway, black] {\texttt{To}} (AA);
\path (A) edge [->, lightpurple, line width=3pt]
node[midway, black] {\texttt{You}} (AB);
\path (A) edge [->, lightpurple, line width=1pt]
node[midway, black] {\texttt{In}} (AC);
\path (A) edge [->, lightpurple, line width=2pt] node[midway, black] {\texttt{<all other tokens>}} (AD);

\node (ACA) at (12,-8) {\ldots};
\node[circle, fill] (ABA) at (4,-8) {};
\node (ABB) at (8,-8) {\ldots};
\node (ABAA) at (12,-12) {\ldots};
\node[circle, fill] (AAA) at (-12,-8) {};
\node (AAB) at (-4,-8) {\ldots};

\path (AC) edge [->, lightpurple, line width=16pt] node[midway, black] {\texttt{<all tokens>}} (ACA);
\path (AB) edge [->, lightpurple, line width=13pt] node[midway, black] {\texttt{can}} (ABA);
\path (ABA) edge [->, lightpurple, line width=16pt] (ABAA);
\path (AB) edge [->, lightpurple, line width=3pt] (ABB);
\path (AA) edge [->, lightpurple, line width=15pt] node[midway, black] {\texttt{copy}} (AAA);
\path (AA) edge [->, lightpurple, line width=1pt] (AAB);

\node[circle, fill] (AAAA) at (-12,-12) {};
\node[circle, fill] (AAAB) at (-4,-12) {};
\node (AAAC) at (4,-12) {\ldots};

\path (AAA) edge [->, lightpurple, line width=5pt] node[midway, black] {\texttt{the}} (AAAA);
\path (AAA) edge [->, lightpurple, line width=2pt] node[midway, black] {\texttt{`/}} (AAAB);
\path (AAA) edge [->, lightpurple, line width=9pt] (AAAC);

\node (Q) at (0,-14) {\vdots};

\node[circle, fill] (QX) at (-12,-16) {};
\node[circle, fill] (X) at (-12,-20) {};
\node[circle, fill] (QY) at (-4,-16) {};
\node[circle, fill] (Y) at (-4,-20) {};
\node[circle, fill] (QZ) at (4,-16) {};
\node[circle, fill] (Z) at (4,-20) {};
\node (W1) at (9,-20) {\ldots};
\node (W2) at (10.5,-20) {\ldots};
\node (W3) at (12,-20) {\ldots};

\path (QX) edge [->, lightpurple, line width=13pt] node[midway, black] {\texttt{\sc End}} (X);
\path (QY) edge [->, lightpurple, line width=10pt] node[midway, black] {\texttt{\sc End}} (Y);
\path (QX) edge [->, lightpurple, line width=3pt] (W1);
\path (QY) edge [->, lightpurple, line width=6pt] (W2);
\path (QZ) edge [->, lightpurple, line width=1pt] (W3);
\path (QZ) edge [->, lightpurple, line width=15pt] node[midway, black] {\texttt{\sc End}} (Z);

\node[circle, fill] (XA) at (-12,-24) {};
\node[circle, fill] (YA) at (-4,-24) {};
\node[circle, fill] (YB) at (4,-24) {};

\node (XB) at (9, -24) {\ldots};
\node (YC) at (10.5, -24) {\ldots};
\node (ZA) at (12, -24) {\ldots};

\path (X) edge [->, lightblue, line width=2pt] (XA);
\path (Y) edge [->, lightblue, line width=2pt] (YA);
\path (Y) edge [->, lightblue, line width=1pt] (YB);

\path (X) edge [->, lightblue, line width=14pt] (XB);
\path (Y) edge [->, lightblue, line width=14pt] (YC);
\path (Z) edge [->, lightblue, line width=15pt] (ZA);

\node (Al) at (0,1) {$E\left(\text{\parbox{164pt}{\texttt{"User: How can I copy `/dir\_1/file` to `/dir\_2`?"}}}\right), \, I$};

\node (AAl) at (-11.7,-4.8) {$E\left(\text{\parbox{99pt}{\texttt{"User: <...> \textbackslash n \, \, \, \, \, \,  Bot: To"}}}\right), \, I$};
\node (ABl) at (-4,-4.8) {$E\left(\text{\parbox{99pt}{\texttt{"User: <...> \textbackslash n \, \, \, \, \, \, Bot: You"}}}\right), \, I$};
\node (ACl) at (4,-4.8) {$E\left(\text{\parbox{99pt}{\texttt{"User: <...> \textbackslash n \, \, \, \, \, \, Bot: In"}}}\right), \, I$};

\node (ABAl) at (4,-8.8) {$E\left(\text{\parbox{99pt}{\texttt{"User: <...> \textbackslash n \, \, \, \, \, \,  Bot: You can"}}}\right), \, I$};
\node (AAAl) at (-12,-8.8) {$E\left(\text{\parbox{99pt}{\texttt{"User: <...> \textbackslash n \, \, \, \, \, \,  Bot: To copy"}}}\right), \, I$};

\node (AAAAl) at (-12,-12.8) {$E\left(\text{\parbox{112pt}{\texttt{"User: <...> \textbackslash n \, \, \, \, \, \,  Bot: To copy the"}}}\right), \, I$};
\node (AAABl) at (-4,-12.8) {$E\left(\text{\parbox{105pt}{\texttt{"User: <...> \textbackslash n \, \, \, \, \, \,  Bot: To copy `/"}}}\right), \, I$};

\node (Xl) at (-13,-20.6) {$E\left(\text{\parbox{222pt}{\texttt{"User: <...> \textbackslash n \, \, \, \, \, \, \, \, \, \, \, \, Bot: To copy the file `/dir\_1/file` to `/dir\_2` <...>"}}}\right), \, I$};
\node (Yl) at (-4,-21.3) {$E\left(\text{\parbox{165pt}{\texttt{"User: <...> \textbackslash n \, \, \, \, \, \,  Bot: To copy `/dir\_1/file` to `/dir\_2`, you can use the `cp` command <...>"}}}\right), \, I$};
\node (Zl) at (4.5,-21.3) {$E\left(\text{\parbox{180pt}{\texttt{"User: <...> \textbackslash n \, \, \, \, \, \,  Bot: You can copy the file `/dir\_1/file` to `/dir\_2` using the `cp` command <...>"}}}\right), \, I $};

\node (XAl) at (-11.8,-25.3) {$E\left(\text{\parbox{222pt}{\texttt{"User: <...> \textbackslash n \, \, \, \, \, \, \, \, \, \, \, \, \, Bot: To copy the file `/dir\_1/file` to `/dir\_2` <...> \textbackslash n \, \, \, \, \, \, \, \, \, \, \, \, User: Thanks!"}}}\right), \, I'$};
\node (YBl) at (-2.6,-25.55) {$E\left(\text{\parbox{165pt}{\texttt{"User: <...> \textbackslash n \, \, \, \, \, \,  Bot: To copy `/dir\_1/file` to `/dir\_2`, you can use the `cp` command <...> \textbackslash n \, \, \, \, \, \,  User: Thanks!"}}}\right), \, I''$};
\node (YAl) at (5.8,-25.55) {$E\left(\text{\parbox{165pt}{\texttt{"User: <...> \textbackslash n \, \, \, \, \, \,  Bot: To copy `/dir\_1/file` to `/dir\_2`, you can use the `cp` command <...> \textbackslash n \, \, \, \, \, \,  User: I tried but <...>"}}}\right), \, I'''$};

\node[draw] (cap1) at (-12.87, 1) {\large \parbox[t][3.7\baselineskip][t]{0.8\linewidth}{For original input \texttt{"User: How can I copy `/dir\_1/file` to `/dir\_2`?"}, the LLM agent starts at an MDP state which corresponds to the embedding $E(\cdot)$ of that input and which may include other information $I$ on the current world state.}};

\node[draw] (cap2) at (-14.27, -1.6) {\large \parbox[t][2.7\baselineskip][t]{0.63\linewidth}{Each transition (red arrow) corresponds to a possible action, i.e.\ token, with width proportional to the probability that the policy selects it;}};

\node[draw] (cap3) at (-17, -5.4) {\large \parbox[t][7.7\baselineskip][t]{0.3\linewidth}{and takes the agent to the state corresponding to the embedding of the new chat history, with the other information $I$ unchanged.
(``\texttt{<...>}'' is text omitted for length.)}};

\node[draw] (cap4) at (-17, -10.5) {\large \parbox[t][1.7\baselineskip][t]{0.3\linewidth}{The LLM keeps \, \, \, \, \, appending tokens \ldots}};

\node[draw] (cap5) at (-16.6, -18) {\large \parbox[t][2.7\baselineskip][t]{0.35\linewidth}{\ldots until it emits the \textsc{End} token, which prints out the full response to the user.}};


\node[draw] (cap6) at (-15.95, -22.85) {\large \parbox[t][1.7\baselineskip][t]{0.43\linewidth}{The next transitions (blue arrows) represent possible replies;}};

\node[draw] (cap7) at (-18.2, -25.3) {\large \parbox[t][3.7\baselineskip][t]{0.158\linewidth}{the other state information may also change.}};

\end{tikzpicture}
}

\caption{Example of constructing an MDP policy from an LLM.
Note that the transitions are split into those depending deterministically on the agent's actions (red) and those depending entirely on the environment (blue).
} \label{fig:llm_agent_construction}
\end{figure}
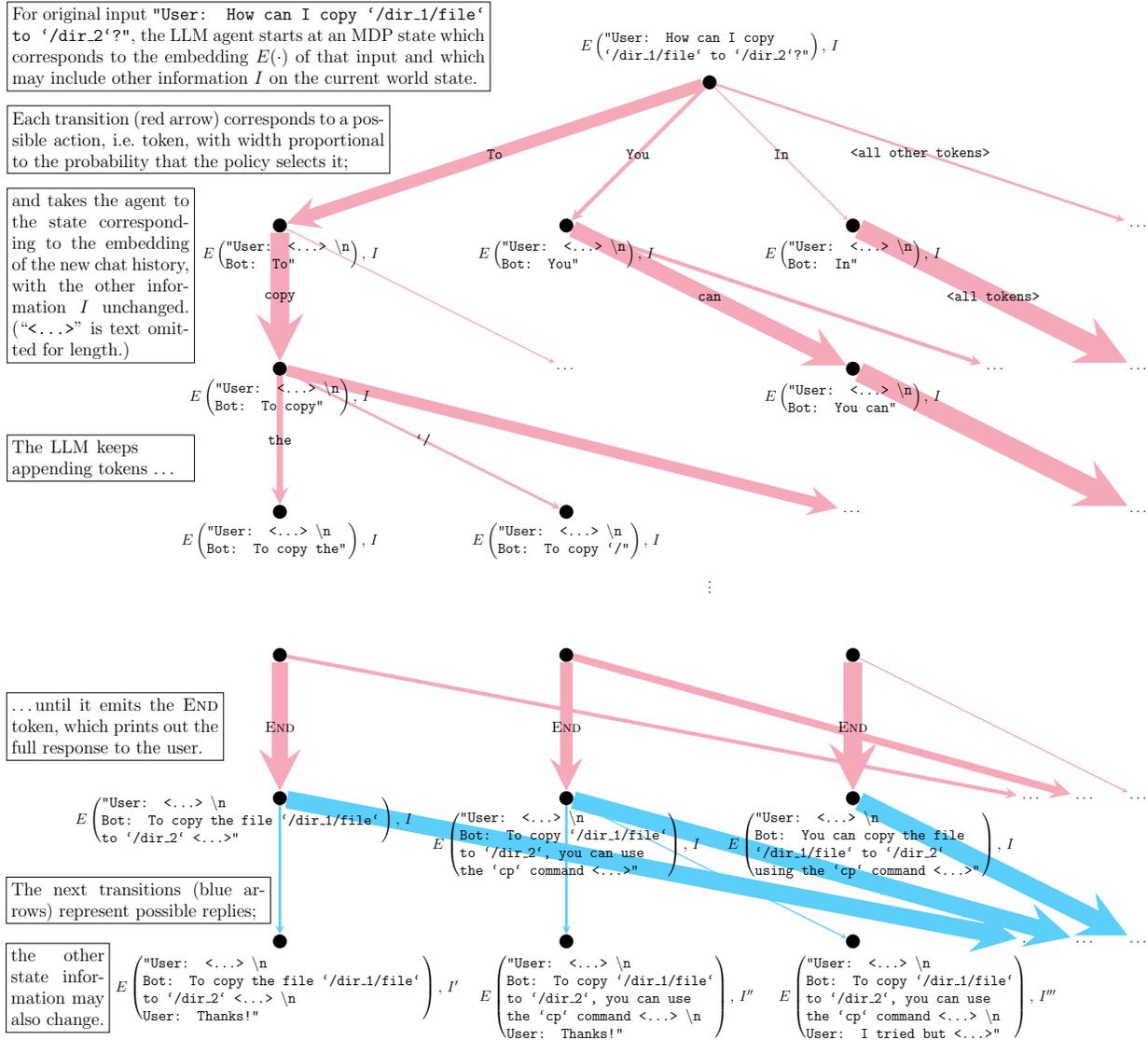


\subsection{Proofs toward Theorem \ref{stability_state_space}} \label{stability_appendix}
First we prove points (\ref{nusc}) and (\ref{lsc}).
\begin{enumerate}
    \item $\mc S_\pi$ is not necessarily upper semicontinuous, because, as in \ref{item:we_solved_alignment_go_home_everyone}, an arbitrarily small change in transition probabilities may suddenly increase the probability of reaching $S_\safe$---e.g.\ for a recurrent state $s_\bad$, the probability of transition $s_\bad \to S_\safe$ changing from 0 to $\epsilon$ redirects all the end-state probability mass from $s_\bad$ to $S_\safe$.
    


    
    \item Note that (\ref{lbrd}) implies (\ref{lsc}) because for any $\eta > 0$, we may take a $\delta < \epsilon, \frac{\eta}{B(\mc M)}$; then $\|\delta \mc M\|_1 < \delta \implies - \frac{\delta \mc S_\pi}{\frac{\eta}{B(\mc M)}} < - \frac{\delta \mc S_\pi}{\|\delta \mc M\|_1} < B(\mc M)$, so $\delta \mc S_\pi > - \eta$, as desired.
    Thus it suffices to prove (\ref{lbrd}).

    \item Our proof of (\ref{lbrd}) will follow this overall structure:
\end{enumerate}

\begin{proofsketch} 
The end behavior of $\mc M$, and hence the value of $\mc S_\pi$, depends on high powers of the transition matrix: $P^N$ as $N \to \infty$.
    To determine the change in the end behavior, then, we do the following:
    \begin{enumerate}
        \item Bound the change in each element of the transition matrix $P$ (Lemma \ref{boundondp}).
        \item Bound $\|\delta P\|_1$, the sum of the absolute values of the changes in the transition matrix, by $\|\delta \mc M\|_1$ (Corollary \ref{dmisdp}).
        \item Express $\mc S_\pi$ in terms of $P$ (Lemma \ref{spiintermsofp}).
        \item Combine these results to obtain a bound on $\frac{|\delta \mc S_\pi|}{\|\delta \mc M\|_1}$ in the case where $S_\trans$, the states from which $S_\safe$ is reachable, is not changed by $\delta \mc M$ (Lemma \ref{boundspinotrans}).
        \item Show that sufficiently small changes in $\mc M$ cannot decrease $|S_\trans|$ (Lemma \ref{localtrans}).
        \item \hyperref[boundspi]{Take the limit as $\|\delta \mc M\|_1 \to 0$ to get a bound on the local rate of decrease, handling the case where $S_\trans$ is changed by $\delta \mc M$.}
    \end{enumerate}
\end{proofsketch}

Now we advance to the full proof.

First, we bound the change in transition probabilities $\delta P$ due to both the change in the environmental transition function (i.e.\ the probabilities of navigating from one state to another given a certain action) and the changes in the policy $\pi$ from the state location changes (i.e.\ the probabilities of selecting a certain action given a current state).

\begin{lemma} \label{boundondp}
    $|\delta P_{ij}| \leq \frac{1}{2} \|\nabla \pi(s_i)\|_1 |\delta s_i| + \sum_a |\delta P(s_i,a,s_j)|$.
\end{lemma}

\begin{proof}
    $\delta P$ depends on $\delta \mc M$ both via changes to the transition probabilities directly and how the changes to the state locations affect the actions chosen by $\pi$.
\begin{align*}
    |\delta P_{ij}|
    = & \Big| \sum_a  \pi(s_i + \delta s_i,a) (P(s_i, a, s_j) + \delta P(s_i,a,s_j)) - \pi(s_i, a) P(s_i, a, s_j) \Big| \\  
    = & \Big| \sum_a \pi(s_i,a) P(s_i, a, s_j) + |\delta s_i| (\nabla_{\delta s_i} \pi(s_i,a)) P(s_i, a, s_j) + \pi(s_i,a) \delta P(s_i,a,s_j) \\
    & + |\delta s_i| (\nabla_{\delta s_i} \pi(s_i,a)) \delta P(s_i,a,s_j) - \pi(s_i, a) P(s_i, a, s_j) \Big| \\
    = & \Big| \sum_a |\delta s_i| (\nabla_{\delta s_i} \pi(s_i,a)) P(s_i, a, s_j) + \pi(s_i,a) \delta P(s_i,a,s_j) + |\delta s_i| (\nabla_{\delta s_i} \pi(s_i,a)) \delta P(s_i,a,s_j) \Big| \\
    = & \Big| \sum_a |\delta s_i| (\nabla_{\delta s_i} \pi(s_i,a)) P(s_i, a, s_j) + \pi(s_i,a) \delta P(s_i,a,s_j) \Big| \hspace{1em}
    \text{removing the $\delta^2$ term.} \\
    \intertext{Note that $|\delta s_i| \nabla_{\delta s_i} \pi(s_i,a)$ is the change in the probability that $\pi$ assigns to action $a$ when $s_i$ moves by $\delta s_i$.
    In other words, this is the contribution from policy change to the change in the transition matrix.
    Since the sum of $\pi$ over all actions must always equal 1, the sum of these changes is 0.
    $0 \leq P(s_i,a,s_j) \leq 1$, so the sum is maximized if $P(s_i,a,s_j)$ is 0 whenever the change is negative and 1 whenever it is positive.
    Since the sum of all the positive changes is half the sum of the absolute value of the changes (again since the changes overall sum to 0), this yields}
    \leq & \frac{1}{2} \sum_a |\delta s_i| |\nabla_{\delta s_i} \pi(s_i, a)| + \Big| \sum_a \pi(s_i,a)\delta P(s_i,a,s_j) \Big|. \\
    \intertext{
    For the first term, note that $\sum_{a} |\nabla_{\delta_{s_i}} \pi(s_i,a)|$ is precisely $\|\nabla_{\delta_{s_i}} \pi(s_i)\|_1$.
    For the second term, note that $\delta P(s_i,a,s_j)$ is the change in the probability of transitioning from $s_i$ to $s_j$ conditional on selecting action $a$.
    Since $\pi(s_i,a)$ is the probability of selecting action $a$, the second sum is the environmental contribution to the change in the transition matrix.
    Since $\pi$ is positive and the sum over all actions of $\pi(s_i,a)$ is 1, we obtain}
    \leq & \frac{1}{2} \|\nabla \pi(s_i)\|_1 |\delta s_i| + \sum_a |\delta P(s_i,a,s_j)|.
\end{align*}
\end{proof}

\begin{corollary} \label{dmisdp}
    $$
    \|\delta P\|_1 \leq \|\delta \mc M\|_1.
    $$
\end{corollary}
\begin{proof}
Recall that
$\|\delta \mc M\|_1
= \frac{1}{2} \cdot |S| \cdot b \cdot \|\delta S\|_1  + \|\delta T\|_1$.
Note that, by definition, $\|\nabla \pi(s_i)\|_1 \leq b$.
    Sum over $i,j$:
$$
\|\delta P\|_1
= \sum_{i,j} |\delta P_{ij}|
\leq \sum_{i,j} \frac{1}{2} \|\nabla \pi(s_i)\|_1 |\delta s_i| + \sum_{i,j,a} |\delta P(s_i,a,s_j)|
\leq \|\delta \mc M\|_1.
$$
\end{proof}
In fact, was the reason we defined $\|\delta \mc M\|_1$ the way we did---the system's behavior is fully described by the transition matrix given by combining $\mc M$'s transition function with a policy, so the size of the change in this transition matrix is a good candidate for a metric on $\mathscr M$.

\begin{lemma} \label{spiintermsofp}
$$
\mc S_\pi = \Delta \cdot \sum_{t=0}^\infty (P I_\trans)^t P v_\safe.
$$
\end{lemma}

\begin{proof}
Let $I_\trans := diag(\mathbb{1}_{s_i \in S_\trans})$ be the matrix that acts as the identity on $S_\trans$, the states of $P$ from which $S_\safe$ is reachable, and which zeroes out all the recurrent states.
Let $v_\safe := \mathbb{1}_{s_i \in S_{rec}}$ be the vector with ones at every safe state and zeroes everywhere else.
Then, $\Delta \cdot (P I_\trans)^t$ is the row vector corresponding to the probability mass in $S_\trans$ at step $t$, noting that we are free to apply the projection $I_\trans$ at every step rather than just at the end because no probability mass can ever return to these states once it has left (by construction).
For convenience, we define $P_\trans = P I_\trans$.
Right-multiplying this by $P$ yields the distribution of the probability mass that was still in $S_\trans$ at step $t$ after one more step.
Finally, taking the inner product with $v_\safe$ sums up the portion of this distribution that is now in $S_\safe$: so each $\Delta \cdot (P_\trans)^t P v_\safe$ is the amount of probability mass that moves into $S_\safe$ at step $t+1$.
So, we can sum over all $t$ to get the total probability mass that moves into $S_\safe$.
\end{proof}

Now we bound the change in $\mc S_\pi$ in terms of the change in $P$.
\begin{lemma} \label{boundspinotrans}
If $\delta \mc M$ does not change $S_\trans$ (i.e.\ for every state, it does not change whether $S_\safe$ is reachable from that state),
    $$
    \frac{|\delta \mc S_\pi|}{\|\delta \mc M\|_1} \leq (1-\lambda_1)^{-1}\left(1 + (1-\lambda_1)^{-1} \right) |S_\safe|,
    $$
    where $\lambda_1$ is the largest eigenvalue of $P_\trans$.
\end{lemma}

\begin{proof}
We express $\delta \mc S_\pi$ in terms of $\delta P$ and $\delta P_\trans$; the latter is well-defined because $\delta \mc M$ does not change $S_\trans$.
\begin{align*}
\delta \mc S_\pi
= & \Delta \cdot \sum_{t=0}^\infty (P_\trans + \delta P_\trans)^t (P + \delta P) v_\safe - \Delta \cdot \sum_{t=0}^\infty P_\trans^t P v_\safe \\
= & \Delta \cdot \left(\sum_{t=0}^\infty (P_\trans + \delta P_\trans)^t (P + \delta P) - \sum_{t=0}^\infty P_\trans^t P \right) v_\safe \\
\intertext{We expand to first order in $\delta P$ using the binomial theorem to obtain}
= & \Delta \cdot \left(\sum_{t=0}^\infty (P_\trans^t + t P_\trans^{t-1} \delta P_\trans) (P + \delta P) -\sum_{t=0}^\infty P_\trans^t P \right) v_\safe \\
\intertext{and expand and once again remove a higher-order $\delta P$ term to get}
= & \Delta \cdot \left(\sum_{t=0}^\infty P_\trans^t P + P_\trans^t \delta P + t P_\trans^{t-1} \delta P_\trans P - \sum_{t=0}^\infty P_\trans^t P \right) v_\safe \\
= & \Delta \cdot \left(\sum_{t=0}^\infty P_\trans^t \delta P + t P_\trans^{t-1} \delta P_\trans P\right)v_\safe \\
= & \Delta \cdot \left(\left(\sum_{t=0}^\infty P_\trans^t \right) \delta P + \left(\sum_{t=0}^\infty t P_\trans^{t-1} \right) \delta P_\trans P\right) v_\safe
\end{align*}
Recall that a geometric series of matrix powers $\sum_{t=0}^\infty A^t$ converges to $(1 - A)^{-1}$ if and only if all the eigenvalues of the matrix are strictly less than 1.
This is indeed the case for $P_\trans$, since only recurrent states have eigenvalue 1.
Also recall that $\sum_{t=1}^\infty t A^{t-1} = (1-A)^{-2}$ since $(1-A) \sum_{t=1}^\infty t A^{t-1} = \sum_{t=1}^\infty t A^{t-1} - \sum_{t=0}^\infty t A^t = \sum_{t=0}^\infty (t+1) A^t - t A^t = \sum_{t=0}^\infty A^t$.
\begin{align*}
\delta \mc S_\pi
= & \Delta \cdot \left(\left(1 - P_\trans\right)^{-1} \delta P + \left(1 - P_\trans\right)^{-2} \delta P I_\trans P\right) v_\safe
\end{align*}
We will bound both of these terms.
Note that the eigenvalues of $(1-P_\trans)^{-1}$ are $\frac{1}{1-\lambda_i}$ where $\lambda_i$ are the eigenvalues of $P_\trans$.
Since $\frac{1}{1-\lambda}$ is monotonically increasing in $\lambda$ for $\lambda < 1$, the largest eigenvalue of $(1-P_\trans)^{-1}$ is $\frac{1}{1-\lambda_1}$, where $\lambda_1$ is the largest eigenvalue of $P_\trans$.
Then,
\begin{align*}
|\delta \mc S_\pi|
\leq & \Delta \cdot \left((1-\lambda_1)^{-1} \|\delta P\|_1 + (1-\lambda_1)^{-2} \|\delta P\|_1 I_\trans P\right) v_\safe \\
\leq & |\Delta| \left((1-\lambda_1)^{-1} \|\delta P\|_1 + (1-\lambda_1)^{-2} \|\delta P\|_1\right) |v_\safe| \\
\leq & (1-\lambda_1)^{-1}\left(1 + (1-\lambda_1)^{-1} \right) |S_\safe| \|\delta \mc M\|_1.
\end{align*}
Dividing by $\|\delta \mc M\|_1$, we obtain the desired inequality.
\end{proof}

Now consider the case where $S_\trans$ is changed by $\delta M$.
\begin{lemma} \label{localtrans}
    For each $\mc M$, $\exists \, \epsilon > 0$ with $\|\delta \mc M\|_1 < \epsilon \implies 
    S_\trans \subset S_\trans'$.
\end{lemma}
\begin{proof}
Consider a state $s$ from which $S_\safe$ was reachable in $\mc M$.
Let $P_s$ be the probability that an agent eventually transitions 
to $S_\safe$ from $s$. By continuity of $P_s$,
for $\|\delta \mc M\|_1 < \epsilon_s$ small enough, $P_s$ will still be positive
in $\mc M'$. Taking the minimum $\epsilon_s$ over all $s$ the result follows.\footnote{The requirement that $\mc M$ and $\mc M'$ be close enough that there are no nonzero transitions of $\mc M$ which become 0 in $\mc M'$ suggests the possibility of using KL divergence to compare the transition functions, since $D_{KL}(P\|Q)$ is infinite if $Q$ lacks support somewhere $P$ does have support.
This may be interesting to apply to scenarios like RLHF on a base model, where the KL divergence is estimated and constrained during training.}
Note that $S_{\trans}'$ may strictly contain $S_{\trans}$: a small perturbation may cause $S_{\safe}$ to be reachable from a state $s$, e.g. by adding a transition directly from $s$ to $S_\safe$.
\end{proof}

\phantomsection \label{boundspi}
Finally, we may conclude the proof of lower semicontinuity with bounded rate of decrease by combining Lemma \ref{boundspinotrans} with Lemma \ref{localtrans}.
Note that if $S'_\trans > S_\trans$, Lemma \ref{boundspinotrans} still provides a lower bound on the safety: if $s' \in S'_\trans \setminus S_\trans$, then $P_\trans + \delta P_\trans$ to excludes entries corresponding to $s'$.
Thus, any probability flow that goes through $s'$ is not included in the bound on safety.
Therefore, the true value of $\mc S_\pi$ accounting for $s'$ can only increase from the lower bound from \ref{boundspinotrans}.
Therefore, we can use Lemma \ref{localtrans} to obtain $\epsilon > 0$ such that
$$
\|\delta \mc M\|_1 < \epsilon
\implies
- \frac{\delta \mc S_\pi}{\|\delta \mc M\|_1}
\leq (1 - \lambda_1)^{-1} \left(1 + (1 - \lambda_1)^{-1}\right) |S_\safe|.
$$

\subsection{Proof of Proposition \ref{delta_is_easy}} \label{proof_delta_easy}

\begin{proof}
    By Lemma \ref{spiintermsofp},
    $\mc S_\pi = \Delta \cdot \sum_{t=0}^\infty (P I_\trans)^t P v_\safe$.
    Since $\mc S_\pi$ is a linear function of $\Delta$, it is continuous in $\Delta$; in fact, it is uniformly continuous:
\begin{align*}
|\mc S_\pi' - \mc S_\pi|
= & \left|(\Delta' - \Delta) \cdot \sum_{t=0}^\infty (P I_\trans)^t P v_\safe\right| \\
\leq & \|\delta \Delta\|_2 \left\|\sum_{t=0}^\infty (P I_\trans)^t P v_\safe\right\|_2 \\
\leq & \|\delta \Delta\|_2,
\end{align*}
which is independent of $\mc M$.
(This also holds for $\|\|_1$ and other reasonable norms for $\delta \Delta$.)
\end{proof}

\section{Discussion: Lower hemicontinuity} \label{appendix: hemicontinuity}
\subsection{Definition}
Hemicontinuity generalizes semicontinuity to set-valued functions.
A lower semicontinuous function must increase only gradually (but may suddenly decrease); a lower hemicontinuous function must move from a point only gradually (but may suddenly add new points).

\begin{definition}
\citep{aliprantis2007}.
A function $\Gamma : A \to 2^B$ is lower hemicontinuous at $a$ if for any open $V \subseteq 2^B$ with $\Gamma(a) \cap V \neq \emptyset$ there is a neighbourhood $U$ of $a$ so that $x \in U \implies \Gamma(x) \cap V \neq \emptyset$.
\end{definition}
The image of $\mc M$ under the functions $\mc F_\delta$ and $\mc F_\pi'$  below represents $\mc M$'s region of safety.
Lower hemicontinuity means that, as $\mc M$ changes, the safe region can recede only gradually, but may suddenly expand to new areas.

\subsection{\texorpdfstring{$\mc F_\delta$}{ℱδ} (defined in section \ref{findings}, item \ref{gen1}) is lower hemicontinuous}

Let $\mc M$ be an MDP and $V \subseteq \mathbb{R}_{\geq 0} \times [0,1]$ be an open set with $\mc F_\delta(\mc M) \cap V \neq \emptyset$.
Let $(N,\varepsilon) \in \mc F_\delta(\mc M) \cap V$.
Since $V$ is an open set, there are some $\eta_1,\eta_2 > 0$ so that $(N+\eta_2,\varepsilon-\eta_1) \in V$.
Now, let $U$ be the ball centered at $\mc M$ with radius $\delta(\eta_1,\eta_2)$ as in remark \ref{remark: stability theorem continuity}.
Then we apply theorem \ref{thm: stability} to find that every $\mc M' \in U$ is $(N + \eta_2, \varepsilon - \eta_1)$-safe, i.e.\ $(N + \eta_2, \varepsilon - \eta_1) \in \mc F_\delta(\mc M')$; thus, $\mc F_\delta(\mc M') \cap V \neq \emptyset$ for all $\mc M' \in U$,
so $\mc F_\delta$ is lower hemicontinuous, as desired.

\subsection{\texorpdfstring{$\mc F_\pi'$}{ℱπ'} (defined in section \ref{findings}, item \ref{gen2}) is lower hemicontinuous}

The proof is extremely similar to that for $\mc F_\delta$.
Let $\mc M$ be an MDP and $V \subseteq \mathbb{P}(S_{\mc M}) \times [0,1]$ be an open set with $\mc F_\pi'(\mc M) \cap V \neq \emptyset$.
Let $(\Delta,p) \in \mc F_\pi'(\mc M) \cap V$.
Since $V$ is an open set, there is some $0 < \eta < p$ so that $(\Delta, p - \eta) \in V$.
Now, let $U$ be the ball centered at $\mc M$ with radius $\min\big\{\epsilon,\frac{\eta}{B(\mc M)}\big\}$, where $\epsilon$ and $B(\mc M)$ are as in theorem \ref{stability_state_space}.
Then we apply theorem \ref{stability_state_space} to find that every $\mc M' \in U$ has $- \frac{\mc S_\pi(\mc M', \Delta) - \mc S_\pi(\mc M, \Delta)}{d_b(\mc M', \mc M)} < B(\mc M)$.
Then $d_b(\mc M',\mc M) < \frac{\eta}{B(\mc M)} \implies - \mc S_\pi(\mc M', \Delta) + \mc S_\pi(\mc M, \Delta) < \eta$
and $p \leq \mc S_\pi(\mc M, \Delta) \implies p - \eta < \mc S_\pi(\mc M', \Delta)$.
Thus $(\Delta, p - \eta) \in \mc F_\pi'(\mc M') \cap V$, so $\mc F_\pi'$ is lower hemicontinuous, as desired.

\end{document}